\documentclass{article}
\usepackage{amsmath, amssymb}
\usepackage{amsthm}
\usepackage{mathtools}
\usepackage{dsfont}
\newtheorem{lemma}{Lemma}
\newtheorem{theorem}{Theorem}
\newtheorem{proposition}{Proposition}
\newtheorem{corollary}{Corollary}
\newtheorem{definition}{Definition}
\newtheorem{assumption}{Assumption}

\usepackage{amsmath}
\usepackage{chngcntr}
\counterwithin{figure}{section}    
\counterwithin{table}{section}     
\counterwithin{theorem}{section}
\counterwithin{definition}{section}
\usepackage[preprint]{neurips_2026}

\usepackage[utf8]{inputenc} 
\usepackage[T1]{fontenc}   
\usepackage{hyperref}     
\usepackage{url}     
\usepackage{booktabs}    
\usepackage{amsfonts}       
\usepackage{nicefrac}     
\usepackage{microtype}  
\usepackage{xcolor} 
\usepackage{algorithm,algpseudocode}
\usepackage{multirow}
\usepackage{makecell}

\definecolor{redorange}{RGB}{255, 68, 51}
\usepackage{color-edits}

\title{Autoregressive Learning in Joint KL:\\
Sharp Oracle Bounds and Lower Bounds}

\author{
  Yunbei Xu\textsuperscript{1}\quad
  Yuzhe Yuan\textsuperscript{1}\quad
  Ruohan Zhan\textsuperscript{2}\\
  \textsuperscript{1}National University of Singapore\quad
  \textsuperscript{2}University College London\\
  \texttt{yunbei@nus.edu.sg}\quad
  \texttt{e1374511@u.nus.edu}\quad
  \texttt{ruohan.zhan@ucl.ac.uk}
}

\definecolor{redorange}{RGB}{255, 68, 51}
\usepackage{color-edits}
\addauthor[Ruohan]{rz}{redorange}

\begin{document}

\maketitle

\begin{abstract}

We study the fundamental and timely problem of learning long sequences in autoregressive
modeling and next-token prediction under model misspecification, measured by
the joint Kullback--Leibler (KL) divergence.  Our goal is to characterize how
the sequence horizon \(H\) affects both approximation and estimation errors in
this joint-distribution, sequence-level regime.  By establishing matching upper and lower
bounds, we provide, to our knowledge, the first complete characterization of
long-horizon error behavior under the natural joint KL objective, with improved
rates and optimality justification relative to existing work.  On the approximation
side, we show that joint KL admits a horizon-free approximation factor, in sharp
contrast to Hellinger-based analyses that exhibit an \(\Omega(H)\) dependence
for computationally efficient methods; this isolates the choice of divergence as
the source of approximation amplification.  On the estimation side, we prove a fundamental
information-theoretic lower bound of order \(\Omega(H)\) that holds for both
decomposable policy classes and fully shared policies, matching the
\(\widetilde O(H)\) upper bounds achieved by computationally efficient
algorithms.  Our analysis clarifies the landscape of recent autoregressive
learning results by aligning the log-loss training objective, the sequence-level
evaluation metric, and the approximation metric  {\color{black}through a sharp joint-KL oracle theory}.  We further show that these joint-KL guarantees imply policy learning
regret bounds  at rates matching 
prior imitation learning literature.

\end{abstract}
  
\section{Introduction}

Learning distributions over long sequences is a central problem in modern AI and in applications of large language models. Autoregressive (AR) modeling addresses this challenge by predicting the next token given the history and repeating this step to generate a full sequence. It does so by factorizing the joint distribution through conditional next-step laws, enabling likelihood-based training and principled probabilistic evaluation. This formulation has become the dominant template for sequence modeling, powering large-scale language models \citep{NIPS2000_728f206c} and naturally extending to multimodal generation \citep{pmlr-v48-oord16}, sequential density estimation \citep{pmlr-v15-larochelle11a}, and model-based forecasting \citep{SALINAS20201181}.

We formalize AR modeling as learning a policy that induces a joint distribution over length-$H$ sequences under the maximum-likelihood objective. Here we adopt the useful dynamical-systems lens
\citep{rohatgi2025computational,hazan2025researchprogramtheorylearning}, to regard the history as a state, the next token as an action or output, and
data-generating mechanism as a policy.
Formally, let $\pi^\star = (\pi_1^\star,\ldots,\pi_H^\star)$ denote the data-generating policy for a length-$H$ sequence $S=(u_1,\ldots,u_H)$, where $\pi_h^\star(\cdot \mid u_{<h})$ is the conditional law of the $h$-th token given its history $u_{<h}$.
Let $P^{\pi^\star}$ be the induced joint distribution over $S$.
For any candidate policy $\pi = (\pi_1,\ldots,\pi_H) \in \Pi$, denote by $P^\pi$ the induced joint distribution.
AR modeling exploits the sequential structure through the autoregressive factorization
\begin{equation}
P^\pi(u_{1:H}) \;=\; \prod_{h=1}^H \pi_h\!\left(u_h \mid u_{<h}\right),
\qquad
u_{<h} = (u_1,\ldots,u_{h-1}).
\end{equation}
The population maximum-likelihood (equivalently, cross-entropy) objective \cite[§20]{goodfellow2016deep}, used explicitly in autoregressive language modeling since GPT-1 \citep{radford2018improving}, aims to minimize the expected negative log-likelihood under $P^{\pi^\star}$ and is therefore equivalent to minimizing the {\it joint} Kullback--Leibler (KL) divergence:
\begin{equation}\label{eq: generative modeling}
\min_{\pi\in\Pi}\; \mathbb{E}_{S\sim P^{\pi^\star}}\!\left[-\sum_{h=1}^H\log \pi_h(u_h|u_{<h})\right]
\;\Longleftrightarrow\;
\min_{\pi\in\Pi}\; \mathrm{KL}\!\left(P^{\pi^\star} \,\|\, P^\pi\right).
\end{equation}
Accordingly, the joint-KL objective \eqref{eq: generative modeling} is the natural target in generative modeling and sequence prediction practice; we refer to learning under this objective as \emph{autoregressive learning}.

Despite its practical importance, autoregressive learning is not a standard
i.i.d. statistical learning problem. The joint-KL objective is
a trajectory-level criterion: later prefixes are generated by earlier tokens, so
the covariate distribution evolves with the horizon rather than being fixed
exogenously. This endogenous sequential structure is the source of the price of long horizons studied in this paper.

In this work, we study the generalization properties of autoregressive learning under {\it misspecification}, where the true policy $\pi^\star$ may not belong to the specified policy class $\Pi$. A standard decomposition of the generalization risk of an estimator $\hat{\pi}$ takes the form
\begin{align*}
    d(P^{\pi^\star},P^{\hat\pi})\leq \underbrace{C_{\text{apx}}\cdot\min\limits_{\pi\in\Pi}d(P^{\pi^\star},P^{\pi})}_{\text{approximation error error}}+\underbrace{C_{\textup{est}}\cdot\frac{\log(|\Pi|/\delta)}{n}+\tilde{\mathcal{O}}(\frac{H}{n})}_{\text{estimation error with fast rate}},
\end{align*}
where $n$ denotes the number of \emph{i.i.d.}~trajectory samples. Throughout, we evaluate models by their \emph{joint} KL to the data-generating process as in \eqref{eq: generative modeling}, which means $d(P,Q)=\textup{KL}(P\|Q)$ and is exactly the population objective of log-loss training.
Unlike standard in-sample conditional density estimation, \eqref{eq: generative modeling}
places autoregressive learning in a distinct joint-distribution regime: the
learned conditional factors are judged by the full sequence law they induce,
so estimation error and misspecification accumulate through the autoregressive
factorization. This raises a basic question: 
{\it under joint sequence-level
metrics, which horizon dependencies in approximation and estimation are intrinsic,
and which are artifacts of the analysis or the metric choices?}

This timely question connects to recent theory on autoregressive learning, next-token
prediction, and in-context learning. Motivated by the maximum-likelihood objective
used in GPTs \citep{radford2018improving}, early studies
\citep{xie2022explanation,wies2023learnability,wang2023large,jiang2023latent}
developed Bayesian or latent-variable interpretations of in-context learning.
These works explain how autoregressive models can perform implicit inference from
context, but they do not provide a misspecified joint-KL oracle theory. Building on this line, \citet{zhang2023and}
formulate next-token prediction as autoregressive density estimation under joint metrics and misspecification, with their main result stated in
joint total variation distance. Although joint-KL-type quantities appear in their
proofs and in-context-learning consequences, their results do not yield a sharp
estimation--approximation decomposition for joint KL metric.
Relatedly, \citet{yuksel2025generalization} obtain KL-type guarantees for
log-loss training of autoregressive processes, but their approximation term is
measured by a worst-case total-variation criterion rather than by the joint KL
metric used for evaluation. None of these works develops lower bounds or an optimality theory
that distinguishes how horizon dependence changes across different structural restrictions on the policy class.

A second line studies autoregressive and imitation learning under squared
Hellinger distance through the minimax-optimality lens. \citet{foster2024behavior} analyze log-loss training for
general behavior cloning and obtain finite-sample guarantees that imply multi-step
control under Hellinger-type evaluation, with autoregressive learning as a special
case. \citet{rohatgi2025computational} focus more directly on misspecified
autoregressive learning and show that, under squared Hellinger evaluation,
computationally tractable algorithms can suffer an \(\Omega(H)\) amplification of
approximation error. These lower bounds are highly informative, but these results also
suggest that the horizon behavior of autoregressive learning is metric-dependent.
Squared Hellinger does not coincide with the population objective induced by
log-loss training, and its pseudo-chain-rule behavior leads to a different
approximation story from the exact KL chain-rule decomposition studied here.
Our work gives a sharp joint-KL analysis in a general autoregressive learning setting.  Both the evaluation risk and the approximation error are measured in the same joint-KL metric, so the result directly characterizes the population objective corresponding to next-token log-loss training.  We provide upper and lower bounds for both approximation and estimation, and treat both fully-shared and decomposable policy classes. Our work provides a unified answer by developing a sharp horizon scaling map for misspecified autoregressive learning under log-loss.
At a high level, Table~\ref{tab:kl-hellinger-summary} summarizes the canonical behaviors of approximation and estimation under joint KL versus squared Hellinger, and clarifies how sharing across steps affects the statistical term.
With this picture in place, we can state the main contributions.

\begin{table}[t]
\centering
\small
\begin{tabular}{l c cc}
\toprule
& \multirow{2}{*}[-0.5ex]{\textbf{Approximation error}}
& \multicolumn{2}{c}{\textbf{Estimation error}} \\
\cmidrule(lr){3-4}
&  & \textbf{Decomposable class} & \textbf{Fully-shared class} \\
\midrule
\textbf{KL divergence (our results)}
& $\Theta(1)$
& $O\!\left(\frac{H\log\!\big(H|\Pi|\big)}{n}\right)$
& $\Theta\!\left(\frac{H\log|\Pi|}{n}\right)$
\\
\makecell[l]{\textbf{Squared Hellinger distance}\\(\cite{rohatgi2025computational})}
& $\Theta(H)$
& $\Theta\!\left(\frac{H\log|\Pi|}{n}\right)$
& $\Theta\!\left(\frac{\log|\Pi|}{n}\right)$
\\
\bottomrule
\end{tabular}
\caption{Summary of scaling behaviors in sequence horizon $H$ and sample size $n$}
\label{tab:kl-hellinger-summary}
\end{table}

\subsection{Main Results and Takeaways}
We summarize our main results and takeaways below; proofs and extensions are deferred to the appendix. 

\begin{itemize}

 \item \textbf{Clarifying the KL formulation in the joint-distribution-level regime.}
We study the new joint-distribution-level regime of autoregressive learning and
organize how prior work has approached this paradigm under TV-, Hellinger-, and
KL-type analyses.  By disentangling the training objective, sequence-level
evaluation metric, and approximation metric, we further clarify the
relationships among existing results and identify joint KL as the
metric-aligned formulation for log-loss training.

 \item {\bf Misspecified guarantees under joint KL (and what drives horizon effects).}
We establish a sharp misspecified generalization guarantee for log-loss, showing that the approximation ratio satisfies $C_{\rm apx}=O(1)$ under the \emph{joint} KL objective.
In particular, in contrast to the $\Omega(H)$ barrier under squared Hellinger distance \citep{rohatgi2025computational}, this shows that horizon growth in $C_{\rm apx}$ is a divergence-induced phenomenon rather than an inherent limitation of autoregressive factorization (see Theorem~\ref{thm:erm-approx}).

\item {\bf Algorithmic upper bounds under joint KL.}
We analyze ERM and Bayesian posterior learners and derive joint-KL generalization bounds with estimation ratio $C_{\rm est}=\tilde{O}(H)$.
The bounds make explicit the tradeoff between approximation and estimation across two canonical policy-class regimes (decomposable versus fully shared), and we extend the analysis to a broad family of dependent policy classes that capture cross-step coupling common in practice (see Section~\ref{sec3.2}). The bounds further imply rollout-level policy learning guarantees to be highlighted below (see also Section~\ref{sec:policy_learning}).

 \item {\bf Information-theoretic lower bounds (unavoidable linear horizon dependence).}
Using a Fano-type construction, we prove that $C_{\rm est}$ must scale at least as $\Omega(H)$ under joint KL (see Theorem~\ref{thm:kl_est_lower_bound}), even for structured classes, showing that the linear horizon dependence in our upper bounds is unavoidable up to logarithmic factors.
This refines the interpretation of ``error amplification'': under log-loss, the fundamental obstruction is statistical rather than approximation-driven.

\end{itemize}

Two points are worth highlighting in more details:

\noindent{\bf Advantages of KL over Hellinger in our setting.}
Our analysis and accompanying insights point to two reasons to favor KL over the squared Hellinger distance adopted in \cite{rohatgi2025computational}. 
First, since {\it  estimation error can be significantly reduced with larger sample size \(n\)}, it is preferable for any horizon dependence \(H\) to appear in the estimation term rather than the approximation term; comparing the first two rows of Table~\ref{tab:kl-hellinger-summary} shows that our KL-based formulation yields the more favorable rate in this regime.
Second, joint KL aligns with the canonical objective in modern generative modeling and large language model training \citep{ermon2023deep,radford2018improving}.

 \noindent{\bf Policy learning guarantee.}
Our joint-KL bounds translate directly into rollout-level policy learning
guarantees: by Pinsker's inequality, they control the worst-case return gap over
bounded trajectory rewards at the usual \(\sqrt{\cdot}\) order.  Hence joint KL
matches the rollout control order obtained from squared Hellinger distance,
while preserving our estimation--approximation separation.

\section{Problem Setting}

Let $\pi^* = (\pi^*_1,\dots, \pi^*_H)$ denote the data-generating  model and $P^{\pi^\star}$ be the induced joint distribution over length-$H$ token sequences.
Given $n$ i.i.d. samples $(S_1, \cdots, S_n)$, where each sequence $S_i\sim P^{\pi^\star}$   consists of tokens $\{u^i_1, \cdots, u_H^i\}$, our goal is to learn a model $\pi$ from a pre-specified class $\Pi$ that best approximates  $\pi^*$ in KL divergence:
\begin{equation}
\min_{\pi \in \Pi} \textup{KL}\!\left(P^{\pi^\star} \,\|\, P^\pi\right).\tag{\ref{eq: generative modeling}}
\end{equation}
In particular, we allow for model misspecification $\pi^*\notin \Pi$. Our objective is therefore to   characterize both the \emph{approximation error} induced by restricting to $\Pi$ and the \emph{estimation error} arising from finite-sample statistical learning. 

Denote by $s_h=u_{<h}$ the length-$(h-1)$ prefix, and write  $\pi=(\pi_1,...,\pi_H)$ for a sequence model. We adopt a widely used assumption of bounded log density \citep{xie2022explanation, rohatgi2025computational}.
\begin{assumption}[Boundedness of log density]\label{assumption}
\begin{align*}
\sup_{\pi\in\Pi}\sup_{h\in [H]}\sup_{s_h}\log \frac{\pi_h^*(x_h|s_h)}{\pi_h(x_h|s_h)}\leq G.
\end{align*}
\end{assumption}

We distinguish between two structural regimes for the policy class $\Pi$.

\noindent{\bf Decomposable policy classes (no parameter sharing).}
In the \emph{decomposable} regime, often called \emph{no parameter sharing}
in prior work (e.g., \citealp{foster2024behavior,rohatgi2025computational})
and motivated by horizon-specific parameterizations practice
(e.g., \citealp{sabbaghi2024explicitly}), the model class factorizes across
time, so each step admits its own independent conditional component. Concretely, the class factorizes across the horizon as
\[
\Pi \;=\; \Pi_1 \times \cdots \times \Pi_H,
\]
where $\Pi_h$ denotes the set of admissible conditional distributions at step $h$.
For ease of presentation, we focus on the homogeneous setting in which the per-step classes coincide,
\[
\Pi_1=\cdots=\Pi_H =: \Pi_0,
\]
so that $\Pi = \Pi_0^H$. Note that heterogeneous per-step classes $(\Pi_h)_{h=1}^H$ can always be reduced to this homogeneous form by taking the lifted base class $
\Pi_0 := \bigcup_{h=1}^H \Pi_h .
$
More generally, one may consider policy classes that do not decompose across steps, i.e., classes satisfying only $\Pi \subseteq \Pi_0^H$. We defer the analysis of such dependent (non-decomposable) classes to Section~\ref{sec3.2}.

\noindent{\bf Fully-shared policy classes (parameter sharing).}
A second regime of primary interest is the \emph{fully shared} 
(or \emph{consistent}) policy class, obtained by tying the same policy component
across all horizons. This restriction is analytically cleaner and is adopted in
prior joint-metric analyses \citep{zhang2023and,yuksel2025generalization}. Formally, this class is given by
\[
\Pi_{\mathrm{cons}}
~:=~
\bigl\{\pi=(\pi_0,\ldots,\pi_0)\in \Pi_0^H:\ \pi_0\in\Pi_0\bigr\},
\]
so that $\pi_h \equiv \pi_0$ for all $h\in[H]$ and $\Pi_{\mathrm{cons}}\subseteq \Pi_0^H$.
This regime captures the canonical parameter-sharing structure in autoregressive model training (e.g., the original Transformer),  while still allowing the induced joint distribution $P^\pi$ to exhibit long-range dependence through the evolving history $u_{<h}$.

\section{Upper Bounds of Autoregressive Learning in Joint KL}
\label{sec:main}
\subsection{Joint-KL Guarantees: Constant Approximation and an Estimation Barrier}
\label{sec:approx}

\noindent{\bf Two canonical log-loss-aligned learners: ERM and Bayesian posterior.}
We begin by studying two classical learners for next-token prediction under the log-loss, both directly aligned with the \emph{joint} KL objective in \eqref{eq: generative modeling}.
The first is empirical risk minimization (ERM), which outputs any minimizer of the empirical negative log-likelihood:
\begin{equation}\label{eq: ERM}
\hat\pi \in \arg\min_{\pi\in\Pi} -\frac{1}{n}\sum_{i=1}^n\sum_{h=1}^H \log \pi_h(u_h^i \mid u_{<h}^i).
\end{equation}
This is the canonical training paradigm for modern language models: Transformer pretraining is precisely maximum-likelihood learning over large text corpora.

As a complementary baseline, we also consider a Bayesian posterior (exponential-weights) learner, which aggregates hypotheses according to their empirical log-likelihood. Concretely, starting from a prior over $\Pi_h$, the per-step posterior after observing $n$ trajectories yields a mixture predictor
\begin{equation}\label{eq: Bayesian}
\hat\pi_h(\cdot\mid s_h)\ \propto\ \sum_{\pi_h\in\Pi_h} q_h(\pi_h)\,\pi_h(\cdot\mid s_h),
\end{equation}
where $q_h$ is the posterior weight updated from the prior by the empirical likelihood $\prod\limits_{i=1}^n\pi_h(u_h^i|u_{<h}^i)$. This learner is also natural under log-loss, in light of the classical mixability viewpoint in online conditional density estimation.

We next show that \emph{both} learners admit oracle-type guarantees under joint KL in the misspecified regime, with horizon-free approximation and horizon-dependent estimation terms. Throughout this paper, we use the notation $ f\lesssim g$ to mean that there exists an absolute constant $c>0$ such that $f\leq c\cdot g$. 

\begin{theorem}[ERM and Bayesian Posterior under joint KL]
\label{thm:erm-approx}
Let $S_1,\dots,S_n \stackrel{\mathrm{iid}}{\sim} P^{\pi^\star}$ with $\pi^\star\notin{\Pi}$, where $\Pi$ is a finite decomposable model class. Under Assumption \ref{assumption}, ERM algorithm \eqref{eq: ERM} and Bayesian Posterior algorithm \eqref{eq: Bayesian} guarantees that for any $\delta\in(0,1)$ and any $\varepsilon>0$, with probability at least $1-\delta$,
\begin{align}\label{eq:erm-approx}
    \textup {KL}(P^{\pi^\star}\|P^{\hat\pi})\lesssim\!\underbrace{(1+\varepsilon)\min\limits_{\pi\in\Pi}\textup {KL}(P^{\pi^\star}\|P^\pi)}_{\text{approximation error}}+\underbrace{\frac{(1+\varepsilon)^2}{\varepsilon}\frac{(HG+1)\log(H|{\Pi_0}|\delta^{-1})}{n}}_{\text{estimation error}}.
\end{align}
For fully shared model class $\Pi_{cons}$, the result would be
\begin{align}\label{eq:erm-approx-cons}
    \textup {KL}(P^{\pi^\star}\|P^{\hat\pi})\lesssim\!\underbrace{(1+\varepsilon)\min\limits_{\pi\in\Pi_{cons}}\textup {KL}(P^{\pi^\star}\|P^\pi)}_{\text{approximation error}}+\underbrace{\frac{(1+\varepsilon)^2}{\varepsilon}\frac{(HG+1)\log(|{\Pi_0}|\delta^{-1})}{n}}_{\text{estimation error}}.
\end{align}
The two regimes trade off approximation and estimation: since $\Pi_{cons}\subset\Pi$, the decomposable class can only reduce the approximation term, while its estimation term differs from the shared case only by an additional $\log H$ factor.
\end{theorem}

The first terms in \eqref{eq:erm-approx} and \eqref{eq:erm-approx-cons} show that the approximation error do not amplify with the horizon (i.e., sequence length): the approximation ratio $C_{\text{apx}}$ on $\min\limits_{\pi\in\Pi}\textup {KL}(P^{\pi^\star}\|P^\pi)$ is  $(1+\varepsilon)$, independent of $H$. 

\noindent{\bf Comparison and technical distinction relative to Hellinger analyses \cite{rohatgi2025computational}.}
The reason why joint KL can achieve horizon independent approximation ratio comes from the \emph{exact chain rule} for KL, which decomposes the joint KL into a sum of conditional KLs along the sequence:
\begin{align*}
\textup{KL}\!\left(P^{\pi^\star},\,P^{\hat\pi}\right)
=\sum_{h=1}^{H}\,
\mathbb{E}_{s_h \sim P^{\pi^\star}(u_{<h})}
\!\Big[\textup{KL}\big(\pi_h^\star(\cdot\mid s_h),\,\hat\pi_h(\cdot\mid s_h)\big)\Big].
\end{align*}
By contrast, the squared Hellinger distance satisfies only a \emph{pseudo chain rule}. A convenient form is 
\setlength{\abovedisplayskip}{4pt}%
\setlength{\belowdisplayskip}{4pt}%
\begin{align*}
\frac{1}{7}\,D_H^2\!\bigl(P^{\pi^\star},P^{\hat\pi}\bigr)
\le
\mathbb{E}_{s_h\sim d_h^{\pi^\star}(u_{<h})}
\Bigl[\sum_{h=1}^{H} D_H^2\!\bigl(\pi_h^\star(\cdot\!\mid\! s_h),\,\hat\pi_h(\cdot\!\mid\! s_h)\bigr)\Bigr]
\le
H\, D_H^2\!\bigl(P^{\pi^\star},P^{\hat\pi}\bigr),
\end{align*}
see, e.g. \citep{foster2024onlineestimationofflineestimation}.
Consequently, stepwise Hellinger analyses incur a linear in $H$ factor, explaining the $\Omega(H)$ dependence reported in prior work \cite{rohatgi2025computational}, whereas the KL chain rule prevents such amplification. This clarifies that the  $\Omega(H)$ growth is a metric-induced artifact.

The proof techniques are also distinct. Their analysis exploits a
sequence-level log-loss--Hellinger relation, together with boundedness and the
triangle inequality, which simplifies dependence analysis but is specific to
Hellinger and leads to the decomposable/shared-class gap reported in
Table~\ref{tab:kl-hellinger-summary}; see the discussion around
\eqref{eq: triangle}. Our joint-KL analysis must instead track prefixes and
union-bound across horizons explicitly. Theorem~\ref{thm:erm-approx} shows that,
even under this sharper metric, decomposable classes incur only \(H\log H\),
rather than the naive \(H^2\), horizon dependence.

\noindent{\bf Improvements over TV and KL analyses \citet{zhang2023and,yuksel2025generalization}.}
\citet{zhang2023and} state their main guarantee in joint total variation.
Since Pinsker converts KL to TV only at square-root scale, a TV guarantee derived from log-loss complexity naturally yields an \(n^{-1/2}\)-type rate,
whereas direct joint-KL analysis can retain the \(n^{-1}\)-type oracle rate
under standard bounded log-loss conditions.

It is also useful to compare Theorem~\ref{thm:erm-approx} with
\citet{yuksel2025generalization}. Their Theorems~3.4 and~3.5 correspond
closely to Step~1 and Step~2 in our proof sketch, but their approximation
term is measured in worst-case total variation, which is not a joint metric:
\[
\mathcal E_{\rm app}
\lesssim
G\cdot \inf_{\pi\in\Pi}\sum_{h=1}^H\sup_{s_h}
\mathrm{TV}\left( \pi^\star_h(\cdot|s_h), \pi_h(\cdot|s_h)\right)
\]
Even strengthening their analysis to sequence-level TV would not give a sharp oracle
bound in the evaluation metric we use, namely joint KL. 
By basic inequalities,  under Assumption \ref{assumption} we have
\[
\mathrm{KL}(P^\star\|P^\pi)
\lesssim
G\cdot \mathbb{E}_{P^{\pi^\star}}\left[\sum_{h=1}^H \mathrm{TV}(\pi_h^{\star}(\cdot|s_h),\pi_h(\cdot|s_h))\right]\lesssim G \sqrt{H \mathrm{KL}(P^\star\|P^\pi)},
\]
showing that even a  sequence-level TV approximation would yield horizon dependence in worst case. 

Moreover, applying the bounds of
\citet{zhang2023and,yuksel2025generalization} to decomposable classes leads
to a naive \(H^2\)-level estimation term. Our ERM
and posterior bounds instead scale linearly in \(H\), again up to logarithms. This advantage is relevant for
horizon- or position-dependent parameterizations used in practice
\citep{sabbaghi2024explicitly}.

\noindent{\bf Core proof ideas.}
The argument first proves a prefix-conditioned oracle inequality for the
one-step log-loss at each fixed step \(h\) and prefix \(s_h\).  For ERM this
uses a Bernstein-type concentration argument, while for the Bayesian posterior
it follows from the mixability of log-loss.  We then average these local
inequalities over the random prefixes generated by the data; a Freedman-type
martingale bound controls the deviation between this empirical prefix average
and its population counterpart.  The exact KL chain rule then converts the
stepwise control into the desired joint-KL bound. In particular, for decomposable classes, the stepwise argument requires
uniform control only over the coordinate-wise index set
\(\{(h,\pi_h):h\in[H],\pi_h\in\Pi_0\}\), yielding the
\(\log(H|\Pi_0|)\) factor instead of the complexity of the full product class, improving the
naive \(H^2\)-horizon dependence to $H\log H$.

\noindent{\bf On the impossibility of $C_{\textup{apx}}=1$ high-probability bounds.}
It is natural to ask whether the Bayesian posterior can satisfy a  high-probability oracle inequality under joint KL with leading constant~$1$ on the misspecification term and only a complexity-dependent remainder. While mixability of log-loss can deliver such sharpness at the prefix-conditioned level (Step~1), it does not extend through the averaging-over-prefixes step needed to control the joint objective (Step~2). As a result, sharp high-probability bounds under joint KL cannot generally be achieved. The following theorem formalizes this impossibility in the misspecified regime, for any learner.
\begin{theorem}[No $C_{\textup{apx}}=1$ high-probability oracle bound under joint KL]
\label{thm:no_sharp_hp}
There exist absolute constants $c>0$ and $\delta_0\in(0,1)$ such that the following holds.
For any $H,n\in\mathbb{N}$, one can construct a misspecified data-generating policy $\pi^\star$
and a base class $\Pi_0$ satisfying Assumption~\ref{assumption} such that for
\[
\Pi\in\Bigl\{\Pi_0^H,\ \Pi_{\mathrm{cons}}:=\{(\pi_0,\ldots,\pi_0):\pi_0\in\Pi_0\}\Bigr\},
\]
every (possibly randomized) learner outputting $\hat\pi\in\Pi$ from $n$ i.i.d.\ trajectories satisfies
\begin{equation}
\Pr\!\left[
\textup{KL}\!\left(P^{\pi^\star}\,\middle\|\,P^{\hat\pi}\right)
\ \ge\
\inf_{\pi\in\Pi}\textup{KL}\!\left(P^{\pi^\star}\,\middle\|\,P^{\pi}\right)
\;+\;
c\,\frac{H}{\sqrt{n}}
\right]
\ \ge\ \delta_0.
\end{equation}
In particular, a sharp high-probability oracle inequality with leading constant $1$
and only a $\widetilde{O}(H\log|\Pi_0|/n)$ remainder is impossible in the misspecified regime.
\end{theorem}

\subsection{Improper Learning via Lifting the Policy Space}\label{sec3.2}

We now turn to the more general \emph{dependent} regime, where the feasible set of stepwise conditionals is coupled across the horizon:
\[
\Pi \subseteq \Pi_1 \times \cdots \times \Pi_H .
\]
Without loss of generality, we take each $\Pi_h$ to be the minimal induced marginal class
\[
\Pi_h := \{\pi_h:\pi \in \Pi\},
\]
so $\Pi_h$ is exactly the collection of step-$h$ conditionals realizable by some $\pi\in\Pi$.
Such dependencies naturally arise from parameter sharing, nonconvex architectural constraints, or combinatorial structures, and they prevent choosing $(\pi_1,\ldots,\pi_H)$ independently. Here in our discussion, ``computational efficiency'' is used in the specific sense of avoiding
the horizon-dependent planning inefficiency highlighted in Section~3 of
\citet{rohatgi2025computational}.

\noindent{\bf Proper learning: coupled search and an $H^2$ statistical barrier.}
A \emph{proper} learner must output a single hypothesis $\hat\pi\in\Pi$.
When $\Pi$ is dependent, this requirement couples all $H$ components, so even evaluating or optimizing the empirical log-loss over $\Pi$
becomes a global structured selection problem over exponentially many joint configurations, which is computationally inefficient in general.
Even if one ignores computation and allows an arbitrary proper selector over $\Pi$, the joint log-loss aggregates along the trajectory and its
effective range and dependence structure can scale with the horizon.
As a consequence, proper learning in dependent classes can suffer an intrinsic estimation barrier of order $\Omega(H^2)$ under joint KL
(see Corollary~\ref{joint-learning} in Appendix), reflecting that long-horizon coupling amplifies statistical uncertainty at the trajectory level.

\noindent{\bf Lifting: tractable stepwise learning with linear horizon scaling.}
We therefore relax properness by \emph{lifting} the search space to the stepwise product class $\tilde\Pi := \Pi_0^H$ where $\Pi_0 := \bigcup_{h=1}^H \Pi_h$ is the marginal union class.
Learning in the lifted space $\tilde\Pi$ enables \emph{stepwise} ERM or Bayesian posterior updates, which are computationally tractable,
and yields an $O(H)$-type estimation term of order $\tilde O(H\log|\Pi_0|/n)$.
The following lemma relates the sizes of $\Pi$ and $\Pi_0$, allowing a direct comparison between the lifted and proper regimes.

\begin{lemma}\label{lifted}
    For any $\Pi\subset\Pi_1\times\cdots\times\Pi_H$ where $\Pi_h:=\{\pi_h:\pi\in\Pi\}$, writing $\Pi_0:=\bigcup_{h=1}^H\Pi_h$, we have that:
    \begin{align*}
    &\log|\Pi_0|\leq\log|\Pi|+\log H\\
    &\log|\Pi|\leq H\log|\Pi_0|
    \end{align*}
\end{lemma}

Therefore, the estimation-error gain from using the lifted class is sandwiched asf
\begin{align*}
    H^2\gtrsim\;\frac{H^2\log|\Pi|}{H\log|\Pi_0|}
\;\gtrsim\;
\frac{H\log|\Pi|}{\log|\Pi|+\log H},
\end{align*}
so the gain is on the order of a factor ranging from $H$ to $H^2$ in the horizon (modulo logarithms), at the cost of learning in the lifted space. We summarize the computational and statistical gain by lifting as the following theorem. 

\begin{theorem}\label{thm:gain_lifting}
With $\tilde\Pi=\Pi_0^H$ as above, stepwise ERM or Bayesian posterior updates are computationally tractable and achieve only
$\tilde O(H\log|\Pi_0|/n)$ estimation, while any proper learner over a dependent class $\Pi$ can incur $\Omega(H^2\log|\Pi|/n)$ estimation under joint KL. By Lemma~\ref{lifted}, optimizing over the lifted space corresponds to an $H$-to-$H^2$ (up to a log term) gap in horizon dependence.
\end{theorem}

\section{Lower Bounds of Autoregressive Learning in Joint KL}
\label{sec:discussion_error_amplification}

In this section we discuss error amplification from both the approximation and estimation perspectives. To facilitate a clean comparison across metrics, we still assume that the candidate policy space is identical across steps and focus on two representative cases: decomposable classes and fully shared classes. In particular, Theorem~\ref{thm:kl_est_lower_bound} provides the paper's main
lower bound. This is a central contribution, as it identifies the intrinsic
horizon-dependent statistical cost of autoregressive learning and clarifies the fundamental
limits of earlier upper-bound analyses.

\noindent{\bf Approximation-side error amplification.}
As discussed earlier, squared-Hellinger analyses imply that any computationally efficient procedure may suffer an approximation ratio scaling as $C_{\mathrm{apx}}=\Omega(H)$ in long-horizon AR modeling. Here we interpret this phenomenon through the lens of \emph{error amplification}. The best-in-class misspecification level
\(
\inf_{\pi\in\Pi} d(P^{\pi^\star},P^\pi)
\)
typically grows with $H$ simply because matching a length-$H$ joint distribution becomes progressively harder. An $\Omega(H)$ inflation in $C_{\mathrm{apx}}$ therefore \emph{amplifies} this horizon-growing approximation gap into a much larger sequence-level error, echoing the classical compounding-error intuition from imitation learning. Our results show that this approximation-side amplification can be \emph{metric-induced}: switching to joint KL removes the $\Omega(H)$ blow-up in the approximation ratio, so the amplification is not an inherent feature of sequential prediction but a consequence of the evaluation criterion.

\noindent{\bf Estimation-side error amplification on KL divergence.}
We now focus on the statistical term and make the dependence on the policy class structure explicit. For squared Hellinger distance, there is a standard guarantee for its estimation error. Here we list the result in the realizable case, same results of estimation error also applied to misspecified case.

\begin{theorem}[\cite{foster2024behavior} Proposition 2.1]
\label{Hellinger-loglossBC}
Fix any expert $\pi^\star\in\Pi$. Let $\hat\pi$ be the output of the ERM algorithm (\ref{eq: ERM}). Then for any $\delta\in(0,1)$, with probability at least $1-\delta$,
\begin{equation}\label{h5}
D_H^2\!\left(P^{\hat\pi},\,P^{\pi^\star}\right)
\;\le\;
2\,\frac{\log\!\big(|\Pi|\,\delta^{-1}\big)}{n}.
\end{equation}
\end{theorem}

This result is particularly informative in our setting because $\log|\Pi|$ \emph{faithfully reflects} whether the class shares parameters across steps. Indeed, if $\Pi=\Pi_0^H$, which represent the no parameter sharing case, then $\log|\Pi| = H\log|\Pi_0|$ and \ref{h5} yields an estimation rate of order $H\log|\Pi_0|/n$. If instead $\Pi$ is fully shared, then $|\Pi|=|\Pi_0|$ and the rate improves to $\log|\Pi_0|/n$. In this sense, squared Hellinger can distinguish whether we are effectively selecting $H$ independent hypotheses or reusing a single hypothesis throughout the horizon.

However, this structural ``discount'' from sharing does \emph{not} persist in the misspecified regime under KL. The main reason is captured by the following lower bound result.

\begin{theorem}
\label{thm:kl_est_lower_bound}
There exist absolute constants $c,c_0>0$ such that for any $|\Pi|\ge 4$, any $n\ge c_0\log|\Pi|$, any horizon $H\in\mathbb{N}$, and any $G\in(0,1]$, one can construct a finite decomposible policy class $\Pi$ and associated distributions $\{P^\pi\}_{\pi\in\Pi}$ on a finite sample space such that for every selector $\hat\pi=\hat\pi(S_{1:n})\in\Pi$ based on $S_1,\ldots,S_n \overset{\mathrm{iid}}{\sim} P^{\pi^\star}$ where $\pi^\star\in\Pi$,
\begin{equation}\label{eq:lb-proper}
\inf_{\hat\pi\in\Pi}\;\sup_{\pi^\star\in\Pi}\;
\mathbb{E}_{P^{\pi^\star}}
\!\left[\mathrm{KL}\!\left(P^{\pi^\star}\,\middle\|\,P^{\hat\pi}\right)\right]
\;\ge\;
c\,\frac{H\,G^2\,\log|\Pi|}{n}.
\end{equation}
\end{theorem}

In particular, although Theorem~\ref{thm:kl_est_lower_bound} is stated under realizability, its implication for misspecification is immediate.
For the decomposable regime, enlarging the oracle family can only enlarge the supremum, hence the minimax risk (and the lower bound) cannot decrease.
Moreover, passing to a fully-shared subclass only \emph{restricts} the learner (shrinks the set in the infimum) and does not constrain the oracle;
under misspecification it is natural that $\pi^\star$ lies in the enlarged decomposable family while violating sharing.
Therefore, under misspecification, KL necessarily retains an $H$-linear estimation dependence in both decomposable and fully-shared cases:
even with full sharing, the statistical term scales as $H\log|\Pi_0|/n$.

The lower bound also explains a key difference between the joint-KL setting and
the squared-Hellinger setting summarized in Table~\ref{tab:kl-hellinger-summary}.  The
distinction is most visible for fully-shared model classes. A central reason is that the estimation term for a
fully-shared class can be horizon-free because the proof exploits a special
log-loss/Hellinger relationship\citep{rohatgi2025computational}.  Roughly, exponential-tail control of the
log-likelihood ratio can be combined with the triangle inequality
\begin{align}\label{eq: triangle}
D_{\mathrm H}^2(P^{\hat\pi},P^{\pi^\star})\le 2D_{\mathrm H}^2(P^{\hat\pi},P^{\bar\pi})+2D_{\mathrm H}^2(P^{\bar\pi},P^{\pi^\star})
\le \frac{4\log(|\Pi|/\delta^{-1})}{n}
+\frac{2}{n}\bigl(\hat L(P^{\bar\pi})-\hat L(P^{\pi^\star})\bigr).
\end{align}
Here \(\widehat L\) denotes the empirical trajectory-level log-loss.  Since this
comparison is made at the level of the induced joint distributions, the
resulting complexity term depends on the shared policy class as a whole, rather
than on a uniform control over horizon-indexed conditional components. This is exactly why their analysis exhibits the {\it striking statistical gap between decomposable and fully shared classes}; see Table~\ref{tab:kl-hellinger-summary} of our paper.

However, this horizon-free behavior is specific to the
log-loss/squared-Hellinger analysis and does not carry over to joint KL.  Under
joint KL, the risk itself decomposes along the autoregressive sequence, so
horizon-wise statistical accumulation cannot be avoided simply by imposing a
fully-shared policy class.  Our lower bound makes this point precise: even in
the fully-shared regime, the estimation error must scale linearly with \(H\).
Thus the \(H\)-dependence in our joint-KL bounds reflects an intrinsic
statistical difficulty of autoregressive learning, rather than an artifact of
using decomposable classes or of a loose horizon-wise union bound.

\noindent{\bf Summary of error amplification under Hellinger and KL analyses.}
To summary, Hellinger and KL exhibit different error amplification behaviors: Hellinger amplifies primarily through the approximation side, while KL shifts the horizon dependence to the estimation side. Since estimation error can be diluted by increasing the sample size $n$, this makes KL a particularly meaningful and practice-aligned metric despite its $H$-linear statistical accumulation.

\section{From Joint KL Guarantees to Policy Learning Regret Bounds}
\label{sec:policy_learning}
Autoregressive generation can also be viewed as a policy-learning problem: a
policy maps each prefix \(s_h\) to a distribution over the next token, and hence induces a distribution \(P^\pi\) over entire rollouts.  This is the same object that appears in imitation learning (more details may refer to \citep{foster2024behavior}), where one compares the rollout distribution of a learned policy with that of an expert policy. To make this connection explicit, let
\[
    J(\pi;r):=\mathbb{E}_{\tau\sim P^\pi}[r(\tau)]
\]
denote the expected return of policy \(\pi\) under a trajectory-level reward
function \(r\).  We call \(r\) \(R\)-bounded if \(|r(\tau)|\le R\) for all
trajectories \(\tau\), and for simplicity take \(R=1\). For the standard imitation-learning formulation, the worst-case, sequence-level regret can be bounded by joint TV as the following:
\[
    \sup_{r:\,1\text{-bounded}}
    \left\{
        J(\pi^\star;r)-J(\widehat\pi;r)
    \right\}
    =
    D_{\mathrm{TV}}(P^{\pi^\star},P^{\widehat\pi}).
\]
Thus controlling the trajectory-level total variation distance is exactly
equivalent to controlling the worst-case return gap over bounded rewards.

Our joint-KL guarantee fits directly into this policy-learning objective.
Indeed, Pinsker's inequality immediately yields a rollout-control guarantee for
the corresponding imitation-learning problem:
\[
    \sup_{r:\,1\text{-bounded}}
    \left\{
        J(\pi^\star;r)-J(\widehat\pi;r)
    \right\}
    \leq
    \sqrt{\frac{1}{2}
    \mathrm{KL}(P^{\pi^\star}\|P^{\widehat\pi})}.
\]

Thus our bound provides a direct rollout-control guarantee for the corresponding
policy-learning problem.  At this level, joint KL yields the same
\(\sqrt{\cdot}\)-order guarantee as squared Hellinger distance, since
trajectory-level Hellinger control is also converted to total variation through
a square-root comparison (See ~Theorem 3.1, \citep{foster2024behavior}).

This perspective clarifies that the joint-KL objective is not merely a
next-token prediction surrogate: it is strong enough to control the rollout
distribution induced by the learned policy.  It also preserves the main
estimation--approximation message of our analysis: approximation does not
amplify with the horizon, while the linear horizon dependence in estimation is intrinsic under misspecification.

\section{Conclusion}
We studied misspecified autoregressive learning through the lens of \emph{joint} KL divergence.
We show that joint KL admits a horizon-free approximation, which implies that the
$\Omega(H)$ approximation blow-up under squared Hellinger can be metric-induced. At the
same time, our analysis reveals a genuine $\Omega(H)$ estimation barrier for proper learners for joint KL, and show that using horizon-dependent decomposable parameterization would not increase the horizon dependence despite logarithms. We further propose an improper Bayesian posterior learner
over a lifted policy space that is computationally tractable. Finally, we extend the autoregressive density estimation results to rollout-level policy learning guarantees.
A natural direction for future work is to use this framework to study LLM-specific settings, including pre-training, in-context learning, and alignment, and to extend our results to transformer architectures and more realistic policy classes.

\bibliography{reference}
\bibliographystyle{apalike}
\appendix

\newpage
\tableofcontents
\section{Proofs of Theorem~\ref{thm:erm-approx} (Main Upper Bound)}

We prove a generalization of Theorem~\ref{thm:erm-approx} that allows for an infinite policy class via a log-ratio covering argument. The finite-class case is recovered by setting $r=0$.

\begin{definition}[Log-ratio cover]\label{def:logcover}
Fix a policy class $\Pi_0$. For $r>0$, we say that $\Pi'\subset\Pi_0$ is an $r$-cover of $\Pi_0$ if for each $\pi\in\Pi_0$, there is some $\pi'\in\Pi'$ with $\log(\pi(x_h|u_h)/\pi'(x_h|u_h))\leq r$ for any $x_h$ and $u_h$. We write $\mathcal{N}_{\log}(\Pi_0,r)$ to denote the cardinality of the smallest $r$-cover of $\Pi_0$.
\end{definition}

The definition here is standard and widely used \citep{foster2021statistical, rohatgi2025computational}.

The proof consists of two ingredients. The first is a \emph{prefix-conditioned control}. Fix a step $h$ and consider the random prefixes $\{s_h^i\}_{i=1}^n$ induced by the trajectories. The goal is to bound the empirical conditional KL of the learner in terms of any conditional comparator $\bar\pi_h$ within the class (up to a complexity term), namely to control
\[
\frac1n\sum_{i=1}^n \textup{KL}\!\bigl(\pi_h^\star(\cdot\mid s_h^i)\,\|\,\hat\pi_h(\cdot\mid s_h^i)\bigr)
\;-\;
(1+\varepsilon)\frac1n\sum_{i=1}^n \textup{KL}\!\bigl(\pi_h^\star(\cdot\mid s_h^i)\,\|\,\bar\pi_h(\cdot\mid s_h^i)\bigr).
\]
For ERM, this bound is proved in Lemma~\ref{thm:erm-approx1}. For the Bayesian posterior learner, the analogous prefix-conditioned bound is proved in Lemma~\ref{thm-bp-approx1}.

The second ingredient is an \emph{averaging over random prefixes} bound, which connects the empirical average over \emph{visited} prefixes to the population quantity in the chain rule. Concretely, the goal is to control the deviation
\[
\sum_{h=1}^H 
\mathbb{E}_{s_h\sim P^{\pi^\star}}
\Bigl[\textup{KL}\!\bigl(\pi_h^\star(\cdot\mid s_h)\,\|\,\pi_h(\cdot\mid s_h)\bigr)\Bigr]
\;-\;
(1+\varepsilon)\frac1n\sum_{i=1}^n\sum_{h=1}^H 
\textup{KL}\!\bigl(\pi_h^\star(\cdot\mid s_h^i)\,\|\,\pi_h(\cdot\mid s_h^i)\bigr),
\]
uniformly over $\pi$ in the policy class. This step is shared by both ERM and the Bayesian posterior and is proved in Lemma~\ref{lem:delta-cover}. Combining the two ingredients and, moreover, using a Freedman-type deviation bound (Lemma~\ref{one-side control}) to control the empirical-to-population gap
\[
\frac{1}{n}\sum_{i=1}^n \textup{KL}\!\bigl(\pi_h^\star(\cdot\mid s_h^i)\,\|\,\bar\pi_h(\cdot\mid s_h^i)\bigr)
\;-\;
(1+\varepsilon)\mathbb{E}_{S\sim P^{\pi^\star}}\textup{KL}\!\bigl(\pi_h^\star(\cdot\mid s_h)\,\|\,\bar\pi_h(\cdot\mid s_h)\bigr),
\]
yields Theorem~\ref{thm:erm-approx}.

\subsection{Proofs for step 1: prefix-conditioned control}

\begin{lemma}[prefix-conditioned control for ERM]
\label{thm:erm-approx1}
Let $S_1,\dots,S_n \stackrel{\mathrm{iid}}{\sim} P^{\pi^\star}$ with $\pi^\star\notin{\Pi}=\Pi_0^H$ or $\Pi_{cons}$, where $\Pi_0$ has a finite log-ratio cover. Under assumption \ref{assumption}, ERM algorithm guarantees that for any $\delta\in(0,1)$ and any $\varepsilon>0$, with probability at least $1-\delta$, for all $\bar\pi_h$:
\begin{align*}
\frac{1}{n}\sum_{i=1}^n\sum_{h=1}^H
\textup{KL}\!\bigl(\pi_h^\star(\cdot\mid s^i_{1:h-1})\,\|\,\hat\pi_h(\cdot\mid s^i_{1:h-1})\bigr)&\leq
(1+\varepsilon)\,
\frac{1}{n}\sum_{i=1}^n\sum_{h=1}^H
\textup{KL}\!\bigl(\pi_h^\star(\cdot\mid s^i_{1:h-1})\,\|\,\bar\pi_h(\cdot\mid s^i_{1:h-1})\bigr)\\&+
\frac{C}{\varepsilon}\,H(G+1)\,
\frac{\log\!\bigl(\mathcal{N}_{\log}(\Pi_0,r)\,\delta^{-1}\bigr)}{n}
+
C' Hr,
\end{align*}
for universal constants $C,C'$.
\end{lemma}

\paragraph{Proof of Lemma~\ref{thm:erm-approx1}:}
Fix $r>0$ and let $\Pi_r\subset\Pi$ be an $r$-cover in the sense of Definition~\ref{def:logcover}.
For each $\pi\in\Pi$, pick a representative $\pi^r\in\Pi_r$ such that for every length-$h$ trajectory $(u_h,s_{1:h-1})$,
\begin{align}\label{Hr}
\Bigl|\log\frac{\pi(u_h\mid s_{1:h-1})}{\pi^r(u_h\mid s_{1:h-1})}\Bigr|
\le r .
\end{align}

We prove the result for a single step and then sum up them. For any $\pi\in\Pi_0$, define
\begin{align*}
Y(\pi;s_{1:h-1})
&:= \frac{1}{H}\sum_{h=1}^H\log\frac{\pi_h^\star(\cdot\mid s_{1:h-1})}{\pi(\cdot\mid s_{1:h-1})},\\
\mu(\pi;s_{1:h-1})
&:= \mathbb{E}_{u_h\sim \pi_h^\star(\cdot\mid s_{1:h-1})}\!\bigl[Y(\pi;s_{1:h-1})\bigr]
= \textup{KL}\!\bigl(\pi_h^\star(\cdot\mid s_{1:h-1})\,\|\,\pi(\cdot\mid s_{1:h-1})\bigr),\\
\hat\mu(\pi;s_{1:h-1})
&:= \frac1n\sum_{i=1}^n Y_i(\pi;s_{1:h-1}).
\end{align*}
For notational convenience, write $\mu_\pi$ for $\mu(\pi;s_{1:h-1})$ and define $\hat\mu_\pi$ analogously.
Let $\bar\pi$ be any fixed policy in the class, and denote by $\bar\pi^r\in\Pi_r$ its representative.

\medskip
We use Lemma~\ref{central to bernstein} to obtain the Bernstein condition.
From $\mathbb{E}_{P^{\pi^\star}}[e^{-Y}]=1$ and $|Y|\le G$, we have $\textup{Var}(Y)\le V(G)\,\mu$ with $V(G)=4+4G$.
Hence, by the two-sided Bernstein inequality, for any $\delta_g\in(0,1)$, with probability at least $1-\delta_g$,
\begin{align}\label{berstein}
|\mu-\hat\mu|
\le
\sqrt{\frac{2V(G)\,\mu\,\log(1/\delta_g)}{n}}
+
\frac{4G}{3}\,\frac{\log(1/\delta_g)}{n}.
\end{align}

Apply \eqref{berstein} simultaneously to all $\pi'\in\Pi_r$ and to
\[
\hat\pi_r \in \arg\min_{\pi'\in\Pi_r}\hat\mu_{\pi'}.
\]
By a union bound with $\delta_g=\delta/|\Pi_r|$, with probability at least $1-\delta$ the following holds for all $\pi'\in\Pi_r$:
\begin{align}\label{ermgap}
\Delta_r(\pi')
\le
\widehat\Delta_r(\pi')
+
a_r\bigl(\sqrt{\mu_{\pi'}}+\sqrt{\mu_{\bar\pi^r}}\bigr)
+
2b_r,
\end{align}
where
\[
\Delta_r(\pi'):=\mu_{\pi'}-\mu_{\bar\pi^r},
\qquad
\widehat\Delta_r(\pi'):=\hat\mu_{\pi'}-\hat\mu_{\bar\pi^r},
\]
and
\[
a_r:=\sqrt{\frac{2V(G)\,\log(|\Pi_r|\delta^{-1})}{n}},
\qquad
b_r:=\frac{4G}{3}\,\frac{\log(|\Pi_r|\delta^{-1})}{n}.
\]

\medskip
By \eqref{Hr}, we have
\begin{align}\label{coverapx}
|Y(\pi)-Y(\pi^r)|\le r
\quad\Rightarrow\quad
|\mu_\pi-\mu_{\pi^r}|\le r,
\qquad
|\hat\mu_\pi-\hat\mu_{\pi^r}|\le r.
\end{align}
Since $\hat\pi$ is an ERM over $\Pi_0$ and $\Pi_r\subset\Pi_0$, \eqref{coverapx} implies
\[
\min_{\pi'\in\Pi_r}\hat\mu_{\pi'}
\ge
\min_{\pi\in\Pi}\hat\mu_\pi - r
=
\hat\mu_{\hat\pi}-r,
\qquad
\hat\mu_{\hat\pi^r}\le \hat\mu_{\hat\pi}+r,
\]
and therefore
\begin{align}\label{ermgap2}
\widehat\Delta_r(\hat\pi^r)
=
\hat\mu_{\hat\pi^r}-\hat\mu_{\bar\pi^r}
\le 2r .
\end{align}

Applying \eqref{ermgap} with $\pi'=\hat\pi^r$ and using
$\mu_{\hat\pi^r}=\Delta_r(\hat\pi^r)+\mu_{\bar\pi^r}$ together with $\sqrt{x+y}\le \sqrt{x}+\sqrt{y}$, we obtain
\begin{align}\label{ermgap3}
\Delta_r(\hat\pi^r)
\le
a_r\bigl(\sqrt{\Delta_r(\hat\pi^r)}+2\sqrt{\mu_{\bar\pi^r}}\bigr)
+
2b_r
+
2r .
\end{align}
Solving $x\le a_r\sqrt{x}+2a_r\sqrt{\mu_{\bar\pi^r}}+2b_r+2r$ yields
\[
\Delta_r(\hat\pi^r)
\le
2a_r^2+4a_r\sqrt{\mu_{\bar\pi^r}}+4b_r+4r .
\]

\medskip
By \eqref{coverapx} and $\Pi_r\subset\Pi_0$, we have $\mu_{\bar\pi}\le \mu_{\bar\pi^r}\le \mu_{\bar\pi}+r$.
Moreover,
\[
\Delta(\hat\pi):=\mu_{\hat\pi}-\mu_{\bar\pi}
\le
\mu_{\hat\pi^r}-\mu_{\bar\pi^r}+2r
=
\Delta_r(\hat\pi^r)+2r .
\]
Combining the previous display with $|\Pi_r|=\mathcal{N}_{\log}(\Pi_0,r)$ gives
\begin{align*}
\Delta(\hat\pi)
\le\;
&4\,\frac{V(G)\,\log\!\bigl(\mathcal{N}_{\log}(\Pi_0,r)\,\delta^{-1}\bigr)}{n}
+
4\sqrt{2}\,
\sqrt{
\frac{V(G)\,\mu_{\bar\pi^r}\,\log\!\bigl(\mathcal{N}_{\log}(\Pi_0,r)\,\delta^{-1}\bigr)}{n}
}\\
&\quad+
\frac{16}{3}\,G\,
\frac{\log\!\bigl(\mathcal{N}_{\log}(\Pi_0,r)\,\delta^{-1}\bigr)}{n}
+
6r .
\end{align*}

Finally, apply $2\sqrt{xy}\le \varepsilon x + \tfrac{y}{\varepsilon}$ to the square-root term with
$x=\mu_{\bar\pi^r}$ and $y=8V(G)\tfrac{\log(\mathcal{N}_{\log}(\Pi_0,r)\delta^{-1})}{n}$,
and use $\mu_{\bar\pi^r}\le \mu_{\bar\pi}+r$.
Thus for any $\varepsilon>0$,
\begin{align*}
\frac{1}{nH}\sum_{i=1}^n
\textup{KL}\!\bigl(\pi^\star(\cdot\mid s^i_{1:h-1})\,\|\,\hat\pi(\cdot\mid s^i_{1:h-1})\bigr)
=
\mu_{\bar\pi}+\Delta(\hat\pi)
\le\;
&(1+\varepsilon)\,
\frac{1}{nH}\sum_{i=1}^n
\textup{KL}\!\bigl(\pi^\star(\cdot\mid s^i_{1:h-1})\,\|\,\bar\pi(\cdot\mid s^i_{1:h-1})\bigr)\\
&\quad+
\frac{C}{\varepsilon}\,(G+1)\,
\frac{\log\!\bigl(\mathcal{N}_{\log}(\Pi_0,r)\,\delta^{-1}\bigr)}{n}
+
C' r,
\end{align*}
\hfill$\square$

\begin{lemma}[prefix-conditioned control for Bayesian Posterior]\label{thm-bp-approx1}
Let $S_1,\dots,S_n \stackrel{\mathrm{iid}}{\sim} P^{\pi^\star}$ with $\pi^\star\notin{\Pi}=\Pi_0^H$ or $\Pi_{cons}$, where $\Pi_0$ has a finite log-ratio cover. Under assumption \ref{assumption}, Bayesian Posterior algorithm guarantees that for any $\varepsilon>0$, we have that,

\begin{align*}
\frac{1}{n}\!\sum_{i=1}^n\!\sum_{h=1}^H
\textup{KL}(\pi_h^\star(\cdot\,|\,s_{1:h-1}^i)\|\hat\pi_h(\cdot\,|\,s_{1:h-1}^i)
&\le
\frac{1}{n}\!\sum_{i=1}^n\!\sum_{h=1}^H
\textup{KL}(\pi_h^\star(\cdot\,|\,s_{1:h-1}^i)\|\bar\pi_h(\cdot\,|\,s_{1:h-1}^i))
\;\notag\\&+\; Hr
\;+\;\frac{H\log \mathcal N_{\log}(\Pi_0,r)}{n}.
\label{eq:empirical-basic-cover}
\end{align*}
\end{lemma}
\paragraph{Proof.}
Let $S_i=(s_1^i,u_1^i,\ldots,s_H^i,u_H^i)$ be i.i.d.\ trajectories from $P^{\pi^\star}$ for $i=1,\ldots,n$.
Fix $r>0$.
Let $\Pi_{h,r}\subset\Pi_h$ be an $r$-cover under the log-ratio metric, i.e., for every
$\pi_h\in\Pi_h$ there exists $\pi_h^r\in\Pi_{h,r}$ such that for all $(u,s_{1:h-1})$,
\[
\Bigl|\log\frac{\pi_h(u\,|\,s_{1:h-1})}{\pi_h^r(u\,|\,s_{1:h-1})}\Bigr|\le r .
\]

Define the per-step Bayesian (exponential-weights) posterior predictor on $\Pi_{h,r}$ with a uniform prior by
\[
\hat\pi_h(\cdot\,|\,s_{1:h-1})=\sum_{\pi_h^r\in\Pi_{h,r}} q_h(\pi_h^r)\,\pi_h^r(\cdot\,|\,s_{1:h-1}),
\qquad
P^{\hat\pi}(S)\;=\;\prod_{h=1}^H \hat\pi_h(u_h\,|\,s_{1:h-1}) .
\]

By log-loss mixability, for any fixed data $S_{1:n}$ and any $\pi_h^r\in\Pi_{h,r}$,
\begin{equation}\label{eq:oneshot-step-cover}
\sum_{i=1}^n \! \bigl[-\log \hat\pi_h(u_h^i\,|\,s_{1:h-1}^i)\bigr]
\;\le\;
\sum_{i=1}^n \! \bigl[-\log \pi_h^r(u_h^i\,|\,s_{1:h-1}^i)\bigr]
\;+\;\log \mathcal N_{\log}(\Pi_h,r).
\end{equation}
Summing \eqref{eq:oneshot-step-cover} over $h=1,\ldots,H$ and adding
$\sum_{i,h}\log \pi_h^\star(u_h^i\,|\,s_{1:h-1}^i)$ to both sides gives
\begin{align}
\frac{1}{n}\!\sum_{i=1}^n\!\sum_{h=1}^H
\log\frac{\pi_h^\star(u_h^i\,|\,s_{1:h-1}^i)}{\hat\pi_h(u_h^i\,|\,s_{1:h-1}^i)}
&\le
\frac{1}{n}\!\sum_{i=1}^n\!\sum_{h=1}^H
\log\frac{\pi_h^\star(u_h^i\,|\,s_{1:h-1}^i)}{\pi_h^r(u_h^i\,|\,s_{1:h-1}^i)}
\;+\;\frac{1}{n}\sum_{h=1}^H \log \mathcal N_{\log}(\Pi_h,r).
\label{eq:empirical-cover}
\end{align}
For any $\pi\in\Pi$ choose its cover representatives $\{\pi_h^r\}_{h=1}^H\subset\Pi_{h,r}$.
By the cover property,
\[
\Biggl|\frac{1}{n}\sum_{i=1}^n\sum_{h=1}^H
\log\frac{\pi_h(u_h^i\,|\,s_{1:h-1}^i)}{\pi_h^r(u_h^i\,|\,s_{1:h-1}^i)}\Biggr|
\;\le\; Hr .
\]
Since (\ref{eq:empirical-cover}) works for any possible sequence, averaging both sides yields
\begin{align}
\frac{1}{n}\!\sum_{i=1}^n\!\sum_{h=1}^H
\textup{KL}(\pi_h^\star(\cdot\,|\,s_{1:h-1}^i)\|\hat\pi_h(\cdot\,|\,s_{1:h-1}^i)
&\le
\frac{1}{n}\!\sum_{i=1}^n\!\sum_{h=1}^H
\textup{KL}(\pi_h^\star(\cdot\,|\,s_{1:h-1}^i)\|\bar\pi_h(\cdot\,|\,s_{1:h-1}^i))
\;\notag\\&+\; Hr
\;+\;\frac{H\log \mathcal N_{\log}(\Pi_0,r)}{n}.
\label{eq:empirical-basic-cover}
\end{align}
\hfill$\square$

\subsection{Proofs for step 2: averaging over random prefixes}

\begin{lemma}\label{lem:delta-cover}
Let $\Pi_0$ be an arbitrary class and let $\Pi=\Pi_0^H$.
Let $S_i=(u_1^i,\ldots,u_H^i)$ be i.i.d.\ trajectories drawn from $P^{\pi^\star}$.
For any $\pi\in\Pi$ and trajectory $S=(u_1,\ldots,u_H)$, define
\begin{equation}\label{eq:delta_def}
\Delta_\pi(S)
:= \textup{KL}(P^{\pi^\star},P^\pi)
-(1+\varepsilon)\sum_{h=1}^H \textup{KL}\!\bigl(\pi_h^\star(\cdot\mid s_h)\,\|\,\pi_h(\cdot\mid s_h)\bigr),
\qquad s_h=(u_1,\ldots,u_{h-1}).
\end{equation}

Fix any $r>0$, and let $\Pi_{0,r}\subset\Pi_0$ be an $r$-cover in the sense of Definition~\ref{def:logcover}, namely for each $\pi_0\in\Pi_0$ there exists $\pi_0^r\in\Pi_{0,r}$ such that for all $(u,s)$,
\[
\Bigl|\log\frac{\pi_0(u\mid s)}{\pi_0^r(u\mid s)}\Bigr|\le r.
\]
Write $\mathcal N_{\log}(\Pi_0,r):=|\Pi_{0,r}|$.
Then for any $\delta\in(0,1)$, with probability at least $1-\delta$, simultaneously for all $\pi\in\Pi$,
\begin{equation}\label{eq:delta_cover_bound}
\frac{1}{n}\sum_{i=1}^n \Delta_\pi(S_i)
\;\le\;
\frac{(1+\varepsilon)^2HG}{\varepsilon n}\log\!\Bigl(\frac{2H\,\mathcal N_{\log}(\Pi_0,r)}{\delta}\Bigr)
\;+\;(1+\varepsilon)Hr .
\end{equation}

For fully shared class $\Pi_{cons}$, we have that for any $\delta\in(0,1)$, with probability at least $1-\delta$, simultaneously for all $\pi\in\Pi_{cons}$,
\begin{equation}\label{eq:delta_cover_bound_full}
\frac{1}{n}\sum_{i=1}^n \Delta_\pi(S_i)
\;\le\;
\frac{(1+\varepsilon)^2HG}{\varepsilon n}\log\!\Bigl(\frac{2\,\mathcal N_{\log}(\Pi_0,r)}{\delta}\Bigr)
\;+\;(1+\varepsilon)Hr .
\end{equation}
In particular, the same bound holds for any data-dependent $\hat\pi\in\Pi$.
\end{lemma}

\paragraph{Proof of Lemma~\ref{lem:delta-cover}:}
We first prove the result for decomposable class. Fix $h\in\{1,\ldots,H\}$ and $\pi_h\in\Pi_0$.
For $i=1,\ldots,n$ define
\[
Y_{h,i}(\pi_h):=\textup{KL}\!\bigl(\pi_h^\star(\cdot\mid S_h^i)\,\|\,\pi_h(\cdot\mid S_h^i)\bigr),
\qquad S_h^i=(u_1^i,\ldots,u_{h-1}^i).
\]
By Assumption~1, $0\le Y_{h,i}(\pi_h)\le G$ almost surely.
Let $\mathcal F_i$ be the natural filtration of $(S_1,\ldots,S_i)$.
Then $Y_{h,i}(\pi_h)$ is $\mathcal F_i$-measurable, and by independence together with the chain rule for KL,
\[
\mathbb{E}_{i-1}\!\bigl[Y_{h,i}(\pi_h)\bigr]
=
\mathbb{E}_{S_h}\textup{KL}\!\bigl(\pi_h^\star(\cdot\mid S_h)\,\|\,\pi_h(\cdot\mid S_h)\bigr),
\]
where $S_h$ denotes a generic prefix distributed according to $P^{\pi^\star}$.

Apply Lemma~\ref{one-side control} to the sequence
$X_i:=Y_{h,i}(\pi_h)\in[0,G]$.
With probability at least $1-\delta_{h,\pi_h}$,
\[
\sum_{i=1}^n \mathbb{E}_{i-1}[Y_{h,i}(\pi_h)]
\le
(1+\varepsilon)\sum_{i=1}^n Y_{h,i}(\pi_h) + \frac{(1+\varepsilon)^2G}{\varepsilon}\log\!\Bigl(\frac{2}{\delta_{h,\pi_h}}\Bigr),
\]
and hence after dividing by $n$ and rearranging,
\begin{equation}\label{eq:delta_basic}
\frac{1}{n}\sum_{i=1}^n\Bigl(\mathbb{E}_{i-1}[Y_{h,i}(\pi_h)]-(1+\varepsilon)Y_{h,i}(\pi_h)\Bigr)
\le
\frac{(1+\varepsilon)^2G}{\varepsilon n}\log\!\Bigl(\frac{2}{\delta_{h,\pi_h}}\Bigr).
\end{equation}

Now apply \eqref{eq:delta_basic} to all $h\in[H]$ and all representatives $\pi_h^r\in\Pi_{0,r}$, and take a union bound with
$\delta_{h,\pi_h^r}=\delta/(H\mathcal N_{\log}(\Pi_0,r))$.
With probability at least $1-\delta$, for all $h$ and all $\pi_h^r\in\Pi_{0,r}$,
\begin{equation}\label{eq:delta_union_cover}
\frac{1}{n}\sum_{i=1}^n\Bigl(\mathbb{E}_{i-1}[Y_{h,i}(\pi_h^r)]-(1+\varepsilon)Y_{h,i}(\pi_h^r)\Bigr)
\le
\frac{(1+\varepsilon)^2G}{\varepsilon n}\log\!\Bigl(\frac{2H\mathcal N_{\log}(\Pi_0,r)}{\delta}\Bigr).
\end{equation}

Fix any $\pi=(\pi_1,\ldots,\pi_H)\in\widetilde{\Pi}$ and let $\pi^r=(\pi_1^r,\ldots,\pi_H^r)$ be its coordinatewise representatives in $\Pi_{0,r}$.
By the log-cover property, for every prefix $s$,
\[
\Bigl|\textup{KL}\!\bigl(\pi_h^\star(\cdot\mid s)\,\|\,\pi_h(\cdot\mid s)\bigr)
-
\textup{KL}\!\bigl(\pi_h^\star(\cdot\mid s)\,\|\,\pi_h^r(\cdot\mid s)\bigr)\Bigr|
=
\Bigl|\mathbb{E}_{u\sim\pi_h^\star(\cdot\mid s)}\log\frac{\pi_h^r(u\mid s)}{\pi_h(u\mid s)}\Bigr|
\le r.
\]
Consequently, for each trajectory $S$,
\[
\Bigl|\sum_{h=1}^H \textup{KL}\!\bigl(\pi_h^\star(\cdot\mid s_h)\,\|\,\pi_h(\cdot\mid s_h)\bigr)
-
\sum_{h=1}^H \textup{KL}\!\bigl(\pi_h^\star(\cdot\mid s_h)\,\|\,\pi_h^r(\cdot\mid s_h)\bigr)\Bigr|
\le Hr.
\]
The same argument applies to the chain-rule expression of the joint KL:
\[
\textup{KL}(P^{\pi^\star},P^\pi)=\sum_{h=1}^H \mathbb{E}_{S_h}\textup{KL}\!\bigl(\pi_h^\star(\cdot\mid S_h)\,\|\,\pi_h(\cdot\mid S_h)\bigr),
\]
hence
\[
\bigl|\textup{KL}(P^{\pi^\star},P^\pi)-\textup{KL}(P^{\pi^\star},P^{\pi^r})\bigr|
\le Hr.
\]
Combining the two displays with the definition \eqref{eq:delta_def} yields
\begin{equation}\label{eq:delta_cover_error}
\Delta_\pi(S)\le \Delta_{\pi^r}(S)+(1+\varepsilon)Hr
\qquad\text{for all trajectories }S.
\end{equation}

Finally, sum \eqref{eq:delta_union_cover} over $h=1,\ldots,H$ and use
$\sum_{h=1}^H \sum_{i=1}^n \mathbb{E}_{i-1}[Y_{h,i}(\pi_h^r)]=n\,\textup{KL}(P^{\pi^\star},P^{\pi^r})$ (chain rule)
to obtain
\[
\frac{1}{n}\sum_{i=1}^n \Delta_{\pi^r}(S_i)
\le
\frac{(1+\varepsilon)^2HG}{\varepsilon}\log\!\Bigl(\frac{2H\mathcal N_{\log}(\Pi_0,r)}{\delta}\Bigr)
\]
Together with \eqref{eq:delta_cover_error}, this proves \eqref{eq:delta_cover_bound}.
Since the bound holds uniformly over $\pi\in\widetilde{\Pi}$, it also holds for any data-dependent $\hat\pi\in\widetilde{\Pi}$.

For fully shared class, just take $\delta_{h,\pi_h^r}=\delta/(\mathcal N_{\log}(\Pi_0,r))$ and the result follows the same steps.\hfill$\square$

\subsection{Proofs of Theorem~\ref{thm:erm-approx}}

\begin{proposition}[Theorem~\ref{thm:erm-approx}, generalization to infinite classes]
\label{thm:erm-approx-infinite}
Let $S_1,\dots,S_n \stackrel{\mathrm{iid}}{\sim} P^{\pi^\star}$ with $\pi^\star\notin{\Pi}$, where $\Pi$ is a finite decomposable model class. Under Assumption \ref{assumption}, ERM algorithm \eqref{eq: ERM} and Bayesian Posterior algorithm \eqref{eq: Bayesian} guarantees that for any $\delta\in(0,1)$ and any $\varepsilon>0$, with probability at least $1-\delta$,
\begin{align}
    \textup {KL}(P^{\pi^\star}\|P^{\hat\pi})\lesssim\!\underbrace{(1+\varepsilon)\min\limits_{\pi\in\Pi}\textup {KL}(P^{\pi^\star}\|P^\pi)}_{\text{approximation error}}+\underbrace{\frac{(1+\varepsilon)^{4/3}}{\varepsilon}\frac{(HG+1)\log(H\mathcal N_{\log}(\Pi_0,r)\delta^{-1})}{n}+(1+\varepsilon)Hr}_{\text{estimation error}}.
\end{align}
For fully shared model class $\Pi_{cons}$, the result would be
\begin{align}
    \textup {KL}(P^{\pi^\star}\|P^{\hat\pi})\lesssim\!\underbrace{(1+\varepsilon)\min\limits_{\pi\in\Pi_{cons}}\textup {KL}(P^{\pi^\star}\|P^\pi)}_{\text{approximation error}}+\underbrace{\frac{(1+\varepsilon)^{4/3}}{\varepsilon}\frac{(HG+1)\log(\mathcal N_{\log}(\Pi_0,r)\delta^{-1})}{n}+(1+\varepsilon)Hr}_{\text{estimation error}}.
\end{align}
\end{proposition}

\paragraph{Proof of Proposition~\ref{thm:erm-approx-infinite}:}
We focus on ERM algorithm under decomposable class, and the proof for the Bayesian posterior or fully shared class are similar. Having had Lemma~\ref{thm:erm-approx1} and Lemma~\ref{lem:delta-cover}, the only term we need to control is
\begin{align}\label{part3}
\frac{1}{n}\sum_{i=1}^n \textup{KL}\!\bigl(\pi_h^\star(\cdot\mid s_h^i)\,\|\,\bar\pi_h(\cdot\mid s_h^i)\bigr)
\;-\;
(1+\varepsilon)\mathbb{E}_{S\sim P^{\pi^\star}}\textup{KL}\!\bigl(\pi_h^\star(\cdot\mid s_h)\,\|\,\bar\pi_h(\cdot\mid s_h)\bigr),
\end{align}

Fix $h\in[H]$ and $r>0$, and let $\Pi_{0,r}\subset\Pi_0$ be an $r$-log-ratio cover .
Pick $\tilde\pi_h\in\Pi_0$ such that
\[
\mathbb{E}_{s_h\sim P^{\pi^\star}}\textup{KL}\!\bigl(\pi_h^\star(\cdot\mid s_h)\,\|\,\tilde\pi_h(\cdot\mid s_h)\bigr)
\le
\inf_{\pi_h\in\Pi_0}\mathbb{E}_{s_h\sim P^{\pi^\star}}\textup{KL}\!\bigl(\pi_h^\star(\cdot\mid s_h)\,\|\,\pi_h(\cdot\mid s_h)\bigr)+r,
\]
which exists by the definition of the infimum.
By the cover property, there is $\tilde\pi_h^r\in\Pi_{0,r}$ such that for all $(u,s)$,
$\bigl|\log(\tilde\pi_h(u\mid s)/\tilde\pi_h^r(u\mid s))\bigr|\le r$, and hence for every prefix $s$,
\[
\Bigl|\textup{KL}\!\bigl(\pi_h^\star(\cdot\mid s)\,\|\,\tilde\pi_h(\cdot\mid s)\bigr)
-\textup{KL}\!\bigl(\pi_h^\star(\cdot\mid s)\,\|\,\tilde\pi_h^r(\cdot\mid s)\bigr)\Bigr|
=
\Bigl|\mathbb{E}_{u\sim\pi_h^\star(\cdot\mid s)}\log\frac{\tilde\pi_h^r(u\mid s)}{\tilde\pi_h(u\mid s)}\Bigr|
\le r.
\]
In particular,
\[
\mathbb{E}_{s_h\sim P^{\pi^\star}}\textup{KL}\!\bigl(\pi_h^\star(\cdot\mid s_h)\,\|\,\tilde\pi_h^r(\cdot\mid s_h)\bigr)
\le
\inf_{\pi_h\in\Pi_0}\mathbb{E}_{s_h\sim P^{\pi^\star}}\textup{KL}\!\bigl(\pi_h^\star(\cdot\mid s_h)\,\|\,\pi_h(\cdot\mid s_h)\bigr)+2r.
\]
Define $Y_i:=\textup{KL}\!\bigl(\pi_h^\star(\cdot\mid s_h^i)\,\|\,\bar\pi_h^r(\cdot\mid s_h^i)\bigr)$ and
$\mathcal{F}_i:=\sigma(S_1,\ldots,S_i)$. By Assumption~1, $0\le Y_i\le G$ a.s., and
$\mathbb{E}_{i-1}[Y_i]=\mathbb{E}_{s_h\sim P^{\pi^\star}}Y$ where
$Y=\textup{KL}\!\bigl(\pi_h^\star(\cdot\mid s_h)\,\|\,\bar\pi_h^r(\cdot\mid s_h)\bigr)$.
Applying Lemma~\ref{one-side control} yields, with probability at least $1-\delta$,
\[
\frac1n\sum_{i=1}^n Y_i
-(1+\varepsilon)\mathbb{E}_{s_h\sim P^{\pi^\star}}Y
\le
\frac{G}{\varepsilon n}\log(\delta^{-1}).
\]
Combining with the choice of $\bar\pi_h^r$ gives (\ref{part3}) with the comparator term written as
$\inf_{\pi_h\in\Pi_0}\mathbb{E}_{s_h\sim P^{\pi^\star}}\textup{KL}(\pi_h^\star\|\pi_h)$, up to an additional $r$-approximation term.
Combining Lemma~\ref{thm:erm-approx1}, Lemma~\ref{lem:delta-cover} and \ref{part3}, we have that 
\begin{align}
    \textup {KL}(P^{\pi^\star}\|P^{\hat\pi})\lesssim\!(1+\varepsilon)^3\min\limits_{\pi\in\Pi}\textup {KL}(P^{\pi^\star}\|P^\pi)+\frac{(1+\varepsilon)^2}{\varepsilon}\frac{(HG+1)\log(H\mathcal N_{\log}(\Pi_0,r)\delta^{-1})}{n}+(1+\varepsilon)^3Hr.
\end{align}

Now the preceding bound is of the form
\[
(1+\varepsilon)^3 A \;+\; \frac{(1+\varepsilon)^2}{\varepsilon}\,B \;+\; (1+\varepsilon)^3Hr,
\]
where $A=\min_{\pi\in\Pi}\textup{KL}(P^{\pi^\star}\|P^\pi)$ and
$B=\frac{(HG+1)\log(H\mathcal N_{\log}(\Pi_0,r)\delta^{-1})}{n}$.
To rewrite the leading factor on $A$ as $(1+\varepsilon)$, set $\tilde\varepsilon:=(1+\varepsilon)^{1/3}-1$ so that
$(1+\tilde\varepsilon)^3=1+\varepsilon$. Substituting $\tilde\varepsilon$ for $\varepsilon$ preserves the $1/\varepsilon$ order in the
estimation term since
\[
\frac{(1+\tilde\varepsilon)^2}{\tilde\varepsilon}
=\frac{(1+\varepsilon)^{2/3}}{(1+\varepsilon)^{1/3}-1}
\lesssim \frac{(1+\varepsilon)^{4/3}}{\varepsilon},
\]
up to universal constants. Consequently, we obtain the equivalent (up to constants) decomposition
\[
\textup {KL}(P^{\pi^\star}\|P^{\hat\pi})
\lesssim\!
\underbrace{(1+\varepsilon)\min\limits_{\pi\in\Pi}\textup {KL}(P^{\pi^\star}\|P^\pi)}_{\textup{approximation error}}
+\underbrace{\frac{(1+\varepsilon)^{4/3}}{\varepsilon}\frac{(HG+1)\log(H\mathcal N_{\log}(\Pi_0,r)\delta^{-1})}{n}+(1+\varepsilon)Hr}_{\textup{estimation error}}.
\]
Following the same proof procedure, the same bound can be stated with an $(1+\varepsilon)$ prefactor in the estimation term for the Bayesian posterior, i.e., replacing $(1+\varepsilon)^{4/3}$ by $(1+\varepsilon)$ (up to universal constants). In the main paper, we state the bounds for both ERM and the Bayesian posterior with the enlarged coefficient $(1+\varepsilon)^2$, since for practical choices of $\varepsilon$ this is just a numerical constant and not a key ingredient of the theorem. \hfill$\square$

\section{Proof of Theorem~\ref{thm:kl_est_lower_bound} (Main Lower Bound)}

Now we turn to the proof overview for Theorem~\ref{thm:kl_est_lower_bound}. We establish the lower bound by an information–theoretic \emph{packing} argument combined with Fano’s inequality.  
We build a small family of policies that are well separated yet individually close to the class average, so the data reveal only limited information about which one is true. Under proper learning the estimator must pick a single policy; when the information is too small, Fano forces a constant chance of picking the wrong one, which yields the desired lower bound combining with the separation. The proof below instantiates this outline.

\paragraph{Proof of Theorem~\ref{thm:kl_est_lower_bound}:}
Let the sample space be $\mathcal{S}=\{1,\ldots,d\}$ with $d\ge |\Pi_0|$. Choose $|\Pi_0|$ vectors
$v^{(1)},\ldots,v^{(|\Pi_0|)}\in\{\pm1\}^{d}$ that are pairwise orthogonal and balanced
(e.g., columns of a Hadamard matrix, padded if needed). For $\varepsilon>0$, define
\begin{align*}
P^{\pi_v}(s)
=\frac{\exp(\varepsilon v_s)}{\sum_{k=1}^{d}\exp(\varepsilon v_k)}
=\frac{e^{\varepsilon v_s}}{d\cosh\varepsilon}\qquad(s\in\mathcal{S}).
\end{align*}
Then, for any $v,w$ and any $s$,
\begin{equation}\tag{$\star$}\label{eq:ratio-star}
    \log\frac{P^{\pi_v}(s)}{P^{\pi_w}(s)}
=\varepsilon\,(v_s-w_s)\in\{-2\varepsilon,0,2\varepsilon\}.
\end{equation}

Taking $\varepsilon\le G/2$ gives the required two–sided ratio bound.

Let the average (uniform) distribution be
\begin{align*}
\bar P(s)=\frac{1}{|\Pi_0|}\sum_{v}P^{\pi_v}(s)=\frac{1}{d}.
\end{align*}
A direct calculation using orthogonality yields, for $v\ne w$,
\begin{align*}
\textup{KL}\!\left(P^{\pi_v}\,\|\,P^{\pi_w}\right)=\varepsilon\tanh\varepsilon,
\qquad
\textup{KL}\!\left(P^{\pi_v}\,\|\,\bar P\right)=\varepsilon\tanh\varepsilon-\log\cosh\varepsilon.
\end{align*}
Using $\tanh x\ge x/2$ and $\log\cosh x\le x^2/2$ for $x\in(0,1]$, we have
\begin{align}
\textup{KL}\!\left(P^{\pi_v}\,\|\,P^{\pi_w}\right)\ge \tfrac12\varepsilon^2,
\qquad
\textup{KL}\!\left(P^{\pi_v}\,\|\,\bar P\right)\le \tfrac12\varepsilon^2. \label{eq:quad-bnds}
\end{align}

Now we construct a hard finite class of $H$-step product policies and apply a per-coordinate Fano argument. Define the $H$-step policy class
\[
\ \Pi:=\Pi_0^H=\Big\{\pi=(\pi_1,\ldots,\pi_H):\ \pi_h\in\Pi_0\ \ \forall h\in[H]\Big\},
\]
where each $\Pi_0$ is the same single-step class indexed by $\{v^{(1)},\ldots,v^{(|\Pi_0|)}\}$. For any index vector
\[
\theta=(\theta_1,\ldots,\theta_H)\in\{1,\ldots,|\Pi_0|\}^H,
\]
let $\pi^\theta\in\Pi$ denote the $H$-step policy with $\pi^\theta_h=\pi_{v^{(\theta_h)}}$, and define the induced distribution on an $H$-step trajectory $S=(S_1,\ldots,S_H)\in\mathcal{S}^H$ by the product form
\begin{align}
P^\theta(S)=\prod_{h=1}^H P^{\pi_{v^{(\theta_h)}}}(S_h).
\label{eq:prod-def}
\end{align}
We observe $n$ i.i.d. trajectories $S^{1},\ldots,S^{n}\sim P^\theta$ and write $S_{1:n}:=(S^{1},\ldots,S^{n})$.

By the chain rule for KL and the product structure \eqref{eq:prod-def}, for any $\theta,\theta'$,
\begin{align}
\textup{KL}\!\left(P^{\theta}\,\|\,P^{\theta'}\right)
=\sum_{h=1}^H \textup{KL}\!\left(P^{\pi_{v^{(\theta_h)}}}\,\|\,P^{\pi_{v^{(\theta'_h)}}}\right)
\ \ge\ \frac{\varepsilon^2}{2}\,d_H(\theta,\theta'),
\label{eq:kl-hamming}
\end{align}
where $d_H(\theta,\theta'):=|\{h:\theta_h\ne \theta'_h\}|$.

Let $\Theta$ be uniform on $\{1,\ldots,|\Pi_0|\}^H$. Consider any estimator $\hat\Theta=\hat\Theta(S_{1:n})\in\{1,\ldots,|\Pi_0|\}^H$. Fix any coordinate $h\in[H]$. Conditional on $\Theta_{-h}:=(\Theta_1,\ldots,\Theta_{h-1},\Theta_{h+1},\ldots,\Theta_H)$, the data in coordinates $\neq h$ are independent of $\Theta_h$ and do not carry information about $\Theta_h$ under the product model \eqref{eq:prod-def}. Therefore,
\begin{align*}
I(\Theta_h;S_{1:n}\mid \Theta_{-h})=I(\Theta_h;S_{1:n,h}),
\end{align*}
where $S_{1:n,h}:=(S_h^1,\ldots,S_h^n)$ are the step-$h$ observations across the $n$ trajectories.

Applying Fano's inequality to the $|\Pi_0|$-ary hypothesis test for $\Theta_h$ yields, for any estimator $\hat\Theta_h$,
\begin{align}
\Pr(\hat\Theta_h\ne \Theta_h)
\ \ge\ 1-\frac{I(\Theta_h;S_{1:n,h})+\log 2}{\log|\Pi_0|}.
\label{eq:fano-h}
\end{align}
Moreover, by the standard mutual information bound via the uniform mixture $\bar P$ and \eqref{eq:quad-bnds},
\begin{align}
I(\Theta_h;S_{1:n,h})
\le \sum_{i=1}^n \max_{v}\textup{KL}\!\left(P^{\pi_v}\,\|\,\bar P\right)
\le \frac{n}{2}\varepsilon^2.
\label{eq:mi-h}
\end{align}
Choose
\begin{align*}
\varepsilon:=\frac{G}{4}\sqrt{\frac{\log|\Pi_0|}{n}},
\end{align*}
and assume $n\ge c_0\,\log|\Pi_0|$ with $c_0$ large enough so that $\varepsilon\le \min\{G/2,1\}$. Then \eqref{eq:fano-h}--\eqref{eq:mi-h} imply, for $|\Pi_0|$ large enough,
\begin{align}
\Pr(\hat\Theta_h\ne \Theta_h)\ \ge\ \frac12
\qquad \text{for all }h\in[H].
\label{eq:err-per-step}
\end{align}
Summing \eqref{eq:err-per-step} over $h$ gives
\begin{align}
\mathbb{E}\big[d_H(\Theta,\hat\Theta)\big]
=\sum_{h=1}^H \Pr(\hat\Theta_h\ne \Theta_h)
\ \ge\ \frac{H}{2}.
\label{eq:ham-exp}
\end{align}

Hence then, using \eqref{eq:kl-hamming} and \eqref{eq:ham-exp}, we have that
\begin{align}
\mathbb{E}\Big[\textup{KL}\!\left(P^{\Theta}\,\|\,P^{\hat\Theta}\right)\Big]
\ \ge\ \frac{\varepsilon^2}{2}\,\mathbb{E}\big[d_H(\Theta,\hat\Theta)\big]
\ \ge\ \frac{H}{4}\,\varepsilon^2.
\label{eq:bayes-risk}
\end{align}
Finally, since $\Theta$ is uniform on $\{1,\ldots,|\Pi_0|\}^H$,
\begin{align*}
\sup_{\theta\in\{1,\ldots,|\Pi_0|\}^H}\ 
\mathbb{E}_{P^\theta}\!\left[\textup{KL}\!\left(P^\theta\,\|\,P^{\hat\Theta}\right)\right]
\ \ge\ 
\mathbb{E}\Big[\textup{KL}\!\left(P^{\Theta}\,\|\,P^{\hat\Theta}\right)\Big],
\end{align*}
and combining with \eqref{eq:bayes-risk} and $\varepsilon^2=\frac{G^2}{16}\frac{\log|\Pi_0|}{n}$ yields
\begin{align*}
\sup_{\theta\in\{1,\ldots,|\Pi_0|\}^H}\ 
\mathbb{E}_{P^\theta}\!\left[\textup{KL}\!\left(P^\theta\,\|\,P^{\hat\Theta}\right)\right]
\ \ge\ 
\frac{H}{4}\cdot \frac{G^2}{16}\cdot \frac{\log|\Pi_0|}{n}
\ =\ 
\frac{1}{64}\,\frac{H\,G^2\log|\Pi_0|}{n}.
\end{align*}
This proves the claim with $c=1/64$. \hfill$\square$

\begin{corollary}\label{joint-learning}
There exist absolute constants $c,c_0>0$ such that for any $|\Pi|\ge 4$, any $n\ge c_0\log|\Pi|$, any horizon $H\in\mathbb{N}$, and any $G\in(0,1]$, one can construct a finite general dependent policy class $\Pi$ and associated distributions $\{P^\pi\}_{\pi\in\Pi}$ on a finite sample space such that for every selector $\hat\pi=\hat\pi(S_{1:n})\in\Pi$ based on $S_1,\ldots,S_n \overset{\mathrm{iid}}{\sim} P^{\pi^\star}$ where $\pi^\star\in\Pi$,
\begin{equation}
\inf_{\hat\pi\in\Pi}\;\sup_{\pi^\star\in\Pi}\;
\mathbb{E}_{P^{\pi^\star}}
\!\left[\mathrm{KL}\!\left(P^{\pi^\star}\,\middle\|\,P^{\hat\pi}\right)\right]
\;\ge\;
c\,\frac{H^2G^2\,\log|\Pi|}{n}.
\end{equation}
\end{corollary}

\paragraph{Proof of Corollary~\ref{joint-learning}:} The proof mirrors Theorem~\ref{thm:kl_est_lower_bound}, but the key change is that under a dependent class the trajectory law no longer decomposes across steps. Hence the \eqref{eq:ratio-star} likelihood-ratio bound must be applied to the \emph{joint} policy (rather than stepwise), which forces the corresponding choice of $\varepsilon$ at the $HG$ scale. With this joint control in place, we apply Fano's inequality directly to the $|\Pi|$-ary testing problem and do not need a per-coordinate Fano argument.

Let the sample space be $\mathcal{S}=\{1,\ldots,d\}$ with $d\ge |\Pi|$. Choose $|\Pi|$ vectors
$v^{(1)},\ldots,v^{(|\Pi|)}\in\{\pm1\}^{d}$ that are pairwise orthogonal and balanced
(e.g., columns of a Hadamard matrix, padded if needed). For $\varepsilon>0$, define
\begin{align*}
P^{\pi_v}(s)
=\frac{\exp(\varepsilon v_s)}{\sum_{k=1}^{d}\exp(\varepsilon v_k)}
=\frac{e^{\varepsilon v_s}}{d\cosh\varepsilon}\qquad(s\in\mathcal{S}).
\end{align*}
Then, for any $v,w$ and any $s$,
\begin{align*}
\log\frac{P^{\pi_v}(s)}{P^{\pi_w}(s)}
=\varepsilon\,(v_s-w_s)\in\{-2\varepsilon,0,2\varepsilon\}.
\end{align*}
Taking $\varepsilon\le HG/2$ gives the required two–sided ratio bound.

Let the average (uniform) distribution be
\begin{align*}
\bar P(s)=\frac{1}{|\Pi|}\sum_{v}P^{\pi_v}(s)=\frac{1}{d}.
\end{align*}
A direct calculation using orthogonality yields, for $v\ne w$,
\begin{align*}
\textup{KL}\!\left(P^{\pi_v}\,\|\,P^{\pi_w}\right)=\varepsilon\tanh\varepsilon,
\qquad
\textup{KL}\!\left(P^{\pi_v}\,\|\,\bar P\right)=\varepsilon\tanh\varepsilon-\log\cosh\varepsilon.
\end{align*}
Using $\tanh x\ge x/2$ and $\log\cosh x\le x^2/2$ for $x\in(0,1]$, we have
\begin{align}
\textup{KL}\!\left(P^{\pi_v}\,\|\,P^{\pi_w}\right)\ge \tfrac12\varepsilon^2,
\qquad
\textup{KL}\!\left(P^{\pi_v}\,\|\,\bar P\right)\le \tfrac12\varepsilon^2. \label{eq:quad-bnds-1}
\end{align}

We use Fano's information inequality here. Let $\pi$ be uniform on $\{1,\ldots,|\Pi|\}$ and observe $S_{1:n}\sim (P^{\pi})^{\otimes n}$.  
For any estimator $\hat \pi$, Fano’s  inequality gives that
\begin{align}
\Pr(\hat \pi\ne \pi)\ \ge\ 1-\frac{I(\pi;S_{1:n})+\log2}{\log |\Pi|}. \label{eq:fano}
\end{align}
By the chain rule and \eqref{eq:quad-bnds-1},
\begin{align}
I(\pi;S_{1:n})
\le \sum_{i=1}^{n}\max_{v}\textup{KL}\!\left(P^{\pi_v}\,\|\,\bar P\right)
\le \frac{n}{2}\varepsilon^{2}. \label{eq:mi}
\end{align}
Choose
\begin{align*}
\varepsilon:=\frac{HG}{4}\sqrt{\frac{\log |\Pi|}{n}},
\end{align*}
and assume $n\ge c_0\log |\Pi|$ so that $\varepsilon\le\min\{HG/2,1\}$.  
Then \eqref{eq:fano}–\eqref{eq:mi} imply, for $|\Pi|$ large enough,
\begin{align}
\Pr(\hat \pi\ne \pi)\ \ge\ 1-\frac{H^2G^2}{32}-\frac{\log2}{\log |\Pi|}\ \ge\ \frac12. \label{eq:err-lb}
\end{align}

Notice that whenever $\hat \pi\ne \pi$, by \eqref{eq:quad-bnds-1},
\begin{align*}
\textup{KL}\!\left(P^{\pi}\,\|\,P^{\pi_{\hat \pi}}\right)\ \ge\ \tfrac12\varepsilon^{2}.
\end{align*}
Therefore,
\begin{align*}
\sup_{\pi^\star\in\Pi}\ \mathbb{E}_{P^{\pi^\star}}
\left[\textup{KL}\!\left(P^{\pi^\star}\,\|\,P^{\hat\pi}\right)\right]
\ \ge\ \Pr(\hat \pi\ne \pi)\cdot \frac{\varepsilon^{2}}{2}
\ \stackrel{\eqref{eq:err-lb}}{\ge}\ \frac{1}{4}\cdot\frac{1}{2}\varepsilon^{2}
\ =\ \frac{1}{64}\,\frac{H^2G^2\log |\Pi|}{n},
\end{align*}
which proves the claim with $c=1/64$. \hfill$\square$

\section{Missing proofs in Section~\ref{sec:main}}
\paragraph{Proof of Theorem~\ref{thm:no_sharp_hp}}
It suffices to prove the claim for the special case $H=1$. Indeed, the general $H$ statement follows by taking a product construction across steps so that the joint $\textup{KL}$ adds over $h$ and the lower bound scales linearly in $H$.

Let the token space be $\mathcal{X}=\{0,1\}$. Fix a constant $a\in(0,1/4)$, and for each $n\ge1$ set
\[
b:=\frac{1}{10\sqrt{n}}.
\]
Consider a base class $\Pi_0=\{\pi^{(+)},\pi^{(-)}\}$ consisting of two (history-independent) conditional models,
\[
\pi^{(+)}(1):=\frac12+a,\qquad \pi^{(-)}(1):=\frac12-a,
\qquad
\pi^{(+)}(0):=1-\pi^{(+)}(1),\quad \pi^{(-)}(0):=1-\pi^{(-)}(1).
\]
For the data-generating policy, define $\pi^\star\notin\Pi_0$ by
\[
\pi^\star(1):=\frac12+b,\qquad \pi^\star(0):=1-\pi^\star(1).
\]
Then $P^{\pi^\star}$ is the induced distribution on $u\in\mathcal{X}$ with $\Pr(u=1)=\frac12+b$, and for any $\pi\in\Pi_0$, $P^\pi$ is the induced distribution with $\Pr(u=1)=\frac12\pm a$. Since $b\neq a$, we have $\pi^\star\notin\Pi_0$, i.e.\ the model is misspecified. Moreover, Assumption~1 holds with absolute constants because all Bernoulli parameters lie in $[\frac12-a,\frac12+a]\subset[1/4,3/4]$, so the log-loss range is uniformly bounded. When $H=1$, both choices in the theorem reduce to the same class $\Pi=\Pi_0$ (since $\Pi_0^H=\Pi_0$ and $\Pi_{\textup{cons}}=\Pi_0$).

A direct computation yields
\[
\textup{KL}\!\bigl(P^{\pi^\star}\|P^{\pi^{(-)}}\bigr)
-\textup{KL}\!\bigl(P^{\pi^\star}\|P^{\pi^{(+)}}\bigr)
=
(2b)\log\frac{\frac12+a}{\frac12-a}.
\]
Choosing $a=1/4$ gives the explicit gap $2b\log 3$. In particular,
\[
\inf_{\pi\in\Pi_0}\textup{KL}\!\bigl(P^{\pi^\star}\|P^\pi\bigr)
=\min\Big\{\textup{KL}\!\bigl(P^{\pi^\star}\|P^{\pi^{(+)}}\bigr),
\textup{KL}\!\bigl(P^{\pi^\star}\|P^{\pi^{(-)}}\bigr)\Big\},
\]
and selecting the ``wrong'' element of $\Pi_0$ increases the \textup{KL} by exactly $2b\log 3$.

Let $\hat\pi$ be any (possibly randomized) learner that outputs $\hat\pi\in\Pi_0$ from $n$ i.i.d.\ samples $u_1,\ldots,u_n\sim P^{\pi^\star}$. Consider the associated test $\varphi:=\mathbf{1}\{\hat\pi=\pi^{(+)}\}$. If instead the data-generating policy is
\[
\pi^\star_{+}(1)=\frac12+b
\qquad\text{or}\qquad
\pi^\star_{-}(1)=\frac12-b,
\]
then Le Cam's inequality gives
\[
\Pr_{P^{\pi^\star_{+}}}(\hat\pi=\pi^{(-)})+\Pr_{P^{\pi^\star_{-}}}(\hat\pi=\pi^{(+)})
\ \ge\ 1-\textup{TV}\bigl((P^{\pi^\star_{+}})^{\otimes n},(P^{\pi^\star_{-}})^{\otimes n}\bigr).
\]
By Pinsker's inequality and tensorization,
\[
\textup{TV}\bigl((P^{\pi^\star_{+}})^{\otimes n},(P^{\pi^\star_{-}})^{\otimes n}\bigr)
\le
\sqrt{\frac12\,\textup{KL}\bigl((P^{\pi^\star_{+}})^{\otimes n}\|(P^{\pi^\star_{-}})^{\otimes n}\bigr)}
=
\sqrt{\frac{n}{2}\textup{KL}\bigl(P^{\pi^\star_{+}}\|P^{\pi^\star_{-}}\bigr)}.
\]
For Bernoulli parameters $\frac12\pm b$, one has the standard bound
\[
\textup{KL}\bigl(P^{\pi^\star_{+}}\|P^{\pi^\star_{-}}\bigr)
\le \frac{(2b)^2}{(\frac12-b)(\frac12+b)}
\le 16b^2,
\]
which implies
\[
\textup{TV}\bigl((P^{\pi^\star_{+}})^{\otimes n},(P^{\pi^\star_{-}})^{\otimes n}\bigr)
\le \sqrt{\frac{n}{2}\cdot 16b^2}
=\frac{\sqrt{8}}{10}
<\frac13.
\]
Therefore,
\[
\Pr_{P^{\pi^\star_{+}}}(\hat\pi=\pi^{(-)})+\Pr_{P^{\pi^\star_{-}}}(\hat\pi=\pi^{(+)})
\ge \frac23,
\]
so for at least one choice of sign $\sigma\in\{+,-\}$ we have
\[
\Pr_{P^{\pi^\star_{\sigma}}}(\hat\pi\neq \pi^{(\sigma)})\ge \frac13.
\]
Fix such a sign and take the corresponding data-generating policy $\pi^\star:=\pi^\star_\sigma$.

Now we can conclude the lower bound. On the event $\{\hat\pi\neq \pi^{(\sigma)}\}$, the excess joint \textup{KL} over the oracle in $\Pi_0$ is exactly $2b\log 3$. Hence, with probability at least $\delta_0:=1/3$,
\[
\textup{KL}\!\bigl(P^{\pi^\star}\|P^{\hat\pi}\bigr)
\ \ge\
\inf_{\pi\in\Pi_0}\textup{KL}\!\bigl(P^{\pi^\star}\|P^\pi\bigr)
\ +\ 2b\log 3
\ =\
\inf_{\pi\in\Pi}\textup{KL}\!\bigl(P^{\pi^\star}\|P^\pi\bigr)
\ +\ c\cdot \frac{1}{\sqrt{n}},
\]
where $c:=\frac{\log 3}{5}$ since $b=1/(10\sqrt{n})$. This proves Theorem~\ref{thm:no_sharp_hp} for $H=1$.

As noted at the beginning, the general $H$ statement follows by taking the product construction across $h\in[H]$, in which case the excess \textup{KL} accumulates additively, yielding the term $cH/\sqrt{n}$ while the same constant-probability event persists.\hfill$\square$

\paragraph{Proof of Lemma~\ref{lifted}:} By definition, $\Pi_0=\bigcup_{h=1}^H \Pi_h$, hence
\[
|\Pi_0|
\le \sum_{h=1}^H |\Pi_h|.
\]
For each $h$, the coordinate projection map $\mathrm{proj}_h:\Pi\to \Pi_h$ given by
$\mathrm{proj}_h(\pi)=\pi_h$ is surjective by definition of $\Pi_h$, and therefore
$|\Pi_h|\le |\Pi|$.
Combining these two bounds yields
\[
|\Pi_0|
\le \sum_{h=1}^H |\Pi_h|
\le \sum_{h=1}^H |\Pi|
= H|\Pi|.
\]
Taking logarithms gives the first inequality.

We next prove the second one. Since $\Pi_h\subseteq \Pi_0$ for all $h$, we have
\[
\Pi \subseteq \Pi_1\times\cdots\times \Pi_H \subseteq \Pi_0^H.
\]
Therefore $|\Pi|\le |\Pi_0^H| = |\Pi_0|^H$. Taking logarithms gives the result.\hfill$\square$

\section{Technical Tools}

\begin{lemma}[Central to Bernstein \citep{pmlr-v54-mehta17a}]\label{central to bernstein}
    Let $X$ be a random variable taking values in $[-B, B]$. Assume that $\mathbb{E}\!\left[e^{-\eta X}\right] \le 1$, then
\begin{align*}
    \mathbb{E}[X^2] \le 4\left(\frac{1}{\eta} + B\right)\mathbb{E}[X].
\end{align*}
\end{lemma}
\begin{lemma}[Freedman's inequality \citep{pmlr-v15-beygelzimer11a}]\label{lem:freedman}
Let $(X_t)_{t\le T}$ be a real-valued martingale difference sequence adapted to a filtration $(\mathcal{F}_t)_{t\le T}$.
If $|X_t|\le R$ almost surely, then for any $\eta\in(0,1/R)$, with probability at least $1-\delta$,
\begin{equation}\label{eq:freedman}
\sum_{t=1}^T X_t
\;\le\;
\eta\sum_{t=1}^T \mathbb{E}_{t-1}\!\left[X_t^2\right]
\;+\;
\frac{\log(\delta^{-1})}{\eta}\,.
\end{equation}
\end{lemma}
The following result is a consequence of Lemma \ref{lem:freedman}.
\begin{lemma}\label{one-side control}
Let $(X_t)_{t \le T}$ be a sequence of random variables adapted to a filtration
$(\mathcal{F}_t)_{t \le T}$. If $0 \le X_t \le R$ almost surely, then with probability
at least $1 - \delta$,
\begin{align}
\sum_{t=1}^T X_t
\;\le\;
(1+\varepsilon)\sum_{t=1}^T \mathbb{E}_{t-1}[X_t]
\;+\;
\frac{R}{\varepsilon}\log(\delta^{-1}),
\label{eq:freedman_1pe_forward}
\end{align}
and also with probability at least $1-\delta$,
\begin{align}
\sum_{t=1}^T \mathbb{E}_{t-1}[X_t]
\;\le\;
(1+\varepsilon)\sum_{t=1}^T X_t
\;+\;
\frac{(1+\varepsilon)^2R}{\varepsilon}\log(\delta^{-1}).
\label{eq:freedman_1pe_reverse}
\end{align}
\end{lemma}

\paragraph{Proof of Lemma~\ref{one-side control}:}
Define the martingale difference sequence
\[
Z_t := X_t-\mathbb{E}_{t-1}[X_t],\qquad t=1,\dots,T.
\]
Since $0\le X_t\le R$, we have $|Z_t|\le R$ almost surely. Moreover,
\[
\mathbb{E}_{t-1}[Z_t^2]
=\textup{Var}_{t-1}(X_t)
\le \mathbb{E}_{t-1}[X_t^2]
\le R\,\mathbb{E}_{t-1}[X_t],
\]
where the last inequality uses $X_t^2\le R X_t$ for $X_t\in[0,R]$.

We apply Freedman's inequality (Lemma~\ref{lem:freedman}) to $(Z_t)_{t\le T}$.
For any $\eta\in(0,1/R)$, with probability at least $1-\delta$,
\begin{align*}
\sum_{t=1}^T Z_t
&\le
\eta\sum_{t=1}^T \mathbb{E}_{t-1}[Z_t^2]
+\frac{\log(\delta^{-1})}{\eta}\\
&\le
\eta R\sum_{t=1}^T \mathbb{E}_{t-1}[X_t]
+\frac{\log(\delta^{-1})}{\eta}.
\end{align*}
Recalling $\sum_{t=1}^T Z_t=\sum_{t=1}^T X_t-\sum_{t=1}^T \mathbb{E}_{t-1}[X_t]$, we obtain
\[
\sum_{t=1}^T X_t
\le
(1+\eta R)\sum_{t=1}^T \mathbb{E}_{t-1}[X_t]
+\frac{\log(\delta^{-1})}{\eta}.
\]
Choosing $\eta=\varepsilon/R$ (which is valid since $\varepsilon\in(0,1)$) yields \eqref{eq:freedman_1pe_forward}.

For \eqref{eq:freedman_1pe_reverse}, apply Freedman's inequality to the martingale difference sequence $-Z_t$.
Equivalently, with probability at least $1-\delta$,
\[
-\sum_{t=1}^T Z_t
\le
\eta\sum_{t=1}^T \mathbb{E}_{t-1}[Z_t^2]
+\frac{\log(\delta^{-1})}{\eta}
\le
\eta R\sum_{t=1}^T \mathbb{E}_{t-1}[X_t]
+\frac{\log(\delta^{-1})}{\eta}.
\]
Since $-\sum_{t=1}^T Z_t=\sum_{t=1}^T \mathbb{E}_{t-1}[X_t]-\sum_{t=1}^T X_t$, this implies
\[
(1-\eta R)\sum_{t=1}^T \mathbb{E}_{t-1}[X_t]
\le
\sum_{t=1}^T X_t+\frac{\log(\delta^{-1})}{\eta}.
\]
Choose $\eta=\frac{\varepsilon}{(1+\varepsilon)R}\in(0,1/R)$, so that $(1-\eta R)^{-1}=1+\varepsilon$. Then
\[
\sum_{t=1}^T \mathbb{E}_{t-1}[X_t]
\le
(1+\varepsilon)\sum_{t=1}^T X_t
+(1+\varepsilon)\frac{\log(\delta^{-1})}{\eta}
=
(1+\varepsilon)\sum_{t=1}^T X_t
+\frac{(1+\varepsilon)^2R}{\varepsilon}\log(\delta^{-1}),
\]
which is \eqref{eq:freedman_1pe_reverse}.\hfill$\square$

\section{Impact Statement and Limitations}\label{sec statement}
\subsection{Broader Impacts}
This paper is primarily theoretical: it studies autoregressive learning under
joint KL evaluation and clarifies how log-loss training, misspecification, and
horizon dependence interact at the sequence-distribution level. Its potential
positive impact is methodological reliability. Many sequential prediction and
decision systems are used through the full trajectories they induce, so evaluating
only conditional accuracy can obscure long-horizon error accumulation.
By aligning the training objective, evaluation metric, and approximation metric
in joint KL, our results provide sharper tools for auditing autoregressive models
and for understanding when full-sequence behavior is controlled.   At the same time,
these results should not be interpreted as deployment guarantees by themselves:
joint-KL control on the training distribution does not ensure validity under
unsupported interventions, distribution shift, hidden confounding, or misaligned
objectives. If sequential data encode historical bias, strategic behavior, or
missing variables, sharper sequence modeling may amplify rather than remove such
risks. Responsible use therefore requires support/identification checks,
robustness analysis, constraint and fairness auditing, and appropriate human
oversight in high-stakes applications.

\subsection{Limitations}
This paper is primarily theoretical. We develop a joint-KL framework for
autoregressive learning, characterize horizon dependence, and connect sequence-level statistical guarantees to downstream decision control. The present work
does not provide large-scale empirical validation or domain-specific deployment
studies. In particular, applying the theory to prediction and decision systems requires
problem-specific modeling choices, sufficient support for the relevant
history-action pairs, and careful treatment of distribution shift and
identification. We view applications to language modeling, dynamical systems, and decision-making as important follow-up directions, and are actively developing empirical and domain-driven extensions of the framework.

\end{document}